\documentclass[journal]{IEEEtran}
\usepackage{amsmath,amsfonts}
\usepackage{algorithmic}
\usepackage{array}
\usepackage[caption=false,font=normalsize,labelfont=sf,textfont=sf]{subfig}
\usepackage{textcomp}
\usepackage{stfloats}
\usepackage{url}
\usepackage{cite}
\usepackage{verbatim}
\usepackage{graphicx}
\usepackage{booktabs}
\usepackage{colortbl} 
\usepackage{xcolor}    
\usepackage{tabularx}
\usepackage{array} 
\usepackage{bm}
\usepackage{multirow}
\usepackage{tikz}
\usepackage[T1]{fontenc}
\usepackage[switch]{lineno}
\usepackage[colorlinks=true, linkcolor=blue, citecolor=blue, urlcolor=blue]{hyperref}
\definecolor{LightBlue}{RGB}{220,230,241}
\def\BibTeX{{\rm B\kern-.05em{\sc i\kern-.025em b}\kern-.08em
    T\kern-.1667em\lower.7ex\hbox{E}\kern-.125emX}}

\newcommand*\circlednum[1]{\tikz[baseline=(char.base)]{
            \node[shape=circle,draw,inner sep=0.5pt] (char) {\fontsize{7}{0}\selectfont #1};}}
\newcommand*\circledlet[1]{\tikz[baseline=(char.base)]{
            \node[shape=circle,draw,inner sep=0.8pt] (char) {\fontsize{8}{0}\selectfont #1};}}
\newcommand*\circledtext[1]{\tikz[baseline=(char.base)]{
            \node[shape=circle,draw,inner sep=0.8pt] (char) {\fontsize{10}{0}\selectfont #1};}}
\begin{document}
\title{Content-Distortion High-Order Interaction \\ for Blind Image Quality Assessment}
\author{\IEEEauthorblockN{Shuai Liu, Qingyu Mao, Chao Li, Jiacong Chen, Fanyang Meng, \\ Yonghong Tian, ~\IEEEmembership{Fellow,~IEEE}, and Yongsheng Liang*, ~\IEEEmembership{Member,~IEEE}}
\thanks{
Shuai Liu, Jiacong Chen and Yongsheng Liang are with the College of Applied Technology, Shenzhen University, Shenzhen 518060, China, also with the College of Big Data and Internet, Shenzhen Technology University, Shenzhen 518118, China (e-mail: liushuai981115@163.com; fscjcong@163.com; liangys@szu.edu.cn). Corresponding author: Yongsheng Liang.

Qingyu Mao is with College of Electronic and Information Engineering, Shenzhen University, Shenzhen
518060, China (e-mail: qingyu.mao@outlook.com).

Chao Li is with the School of Electronics and Information Engineering, Harbin Institute of Technology, Shenzhen 518055, China (e-mail: lcc2332021@163.com). 

Fanyang Meng and Yonghong Tian are with the Peng Cheng Laboratory, Shenzhen
518055, China (e-mail: mengfy@pcl.ac.cn; tianyh@pcl.ac.cn).}}

\maketitle
\begin{abstract}
The content and distortion are widely recognized as the two primary factors affecting the visual quality of an image. While existing No-Reference Image Quality Assessment (NR-IQA) methods have modeled these factors, they fail to capture the complex interactions between content and distortions. This shortfall impairs their ability to accurately perceive quality. To confront this, we analyze the key properties required for interaction modeling and propose a robust NR-IQA approach termed CoDI-IQA (Content-Distortion high-order Interaction for NR-IQA), which aggregates local distortion and global content features within a hierarchical interaction framework. Specifically, a Progressive Perception Interaction Module (PPIM) is proposed to explicitly simulate how content and distortions independently and jointly influence image quality. By integrating internal interaction, coarse interaction, and fine interaction, it achieves high-order interaction modeling that allows the model to properly represent the underlying interaction patterns. To ensure sufficient interaction, multiple PPIMs are employed to hierarchically fuse multi-level content and distortion features at different granularities. We also tailor a training strategy suited for CoDI-IQA to maintain interaction stability. Extensive experiments demonstrate that the proposed method notably outperforms the state-of-the-art methods in terms of prediction accuracy, data efficiency, and generalization ability.
\end{abstract}


\section{Introduction}
\label{sec:1}
\IEEEPARstart{I}{mage} quality assessment (IQA) aims to develop objective quality metrics that align with human visual perception \cite{lee2002aimq}. A reliable IQA method is crucial for social media platforms to monitor visual content quality, ensuring a superior visual experience for users \cite{saha2023ReIQA}. Additionally, it can be used as a testing benchmark or optimization goal for image processing algorithms \cite{ding2021Cfr}. Depending on the availability of reference images, IQA can be classified into Full-Reference IQA (FR-IQA), Reduced-Reference IQA (RR-IQA), and No-Reference IQA (NR-IQA) or Blind IQA (BIQA). In real-world scenarios, NR-IQA methods are more applicable as they do not require reference images for evaluation.
\begin{figure}[t]
\centering
\includegraphics[width=0.40\textwidth]{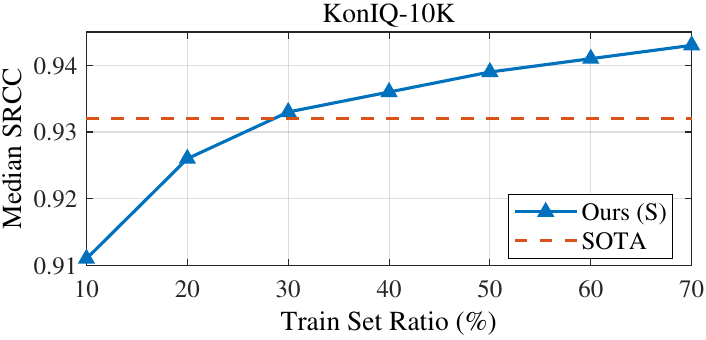}
\includegraphics[width=0.42\textwidth]{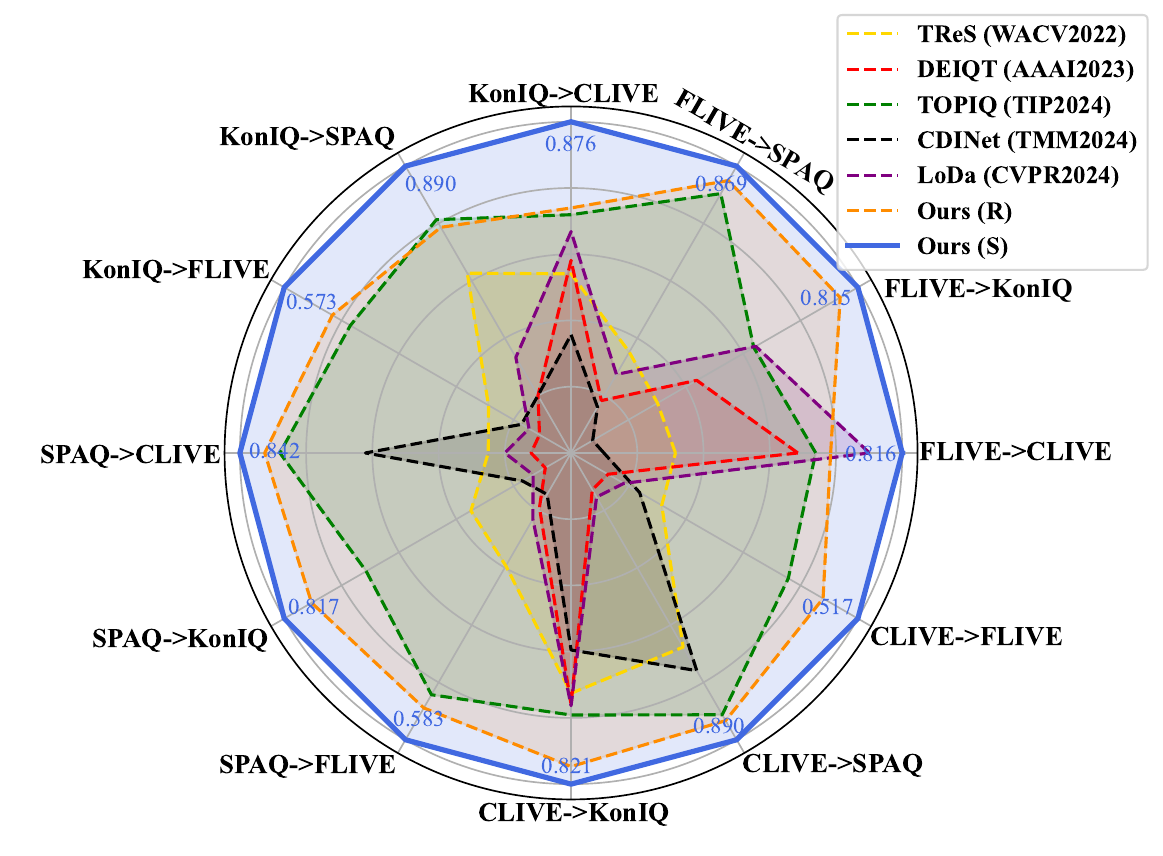}
\caption{Image on top: performance of the proposed CoDI-IQA with varying amounts of training data on the KonIQ-10K \cite{KonIQ} dataset. The state-of-the-art (SOTA) results are obtained from LoDa \cite{xu2024LoDa} with 80\% data, whereas CoDI-IQA can outperforms it using only 30\% data. Image at bottom: Ours CoDI-IQA compared with several SOTA models, showing exceptional improvements in cross-dataset settings on real-world images. The evaluation metric used here is SRCC. Ours (R) and Ours (S) denote CoDI-IQA using ResNet50 \cite{he2016resnet} and Swin-Base Transformer \cite{liu2021swin} as the CAE, respectively.}
\label{fig1}
\end{figure}

\begin{figure}[t]
\centering
\includegraphics[width=1\columnwidth]{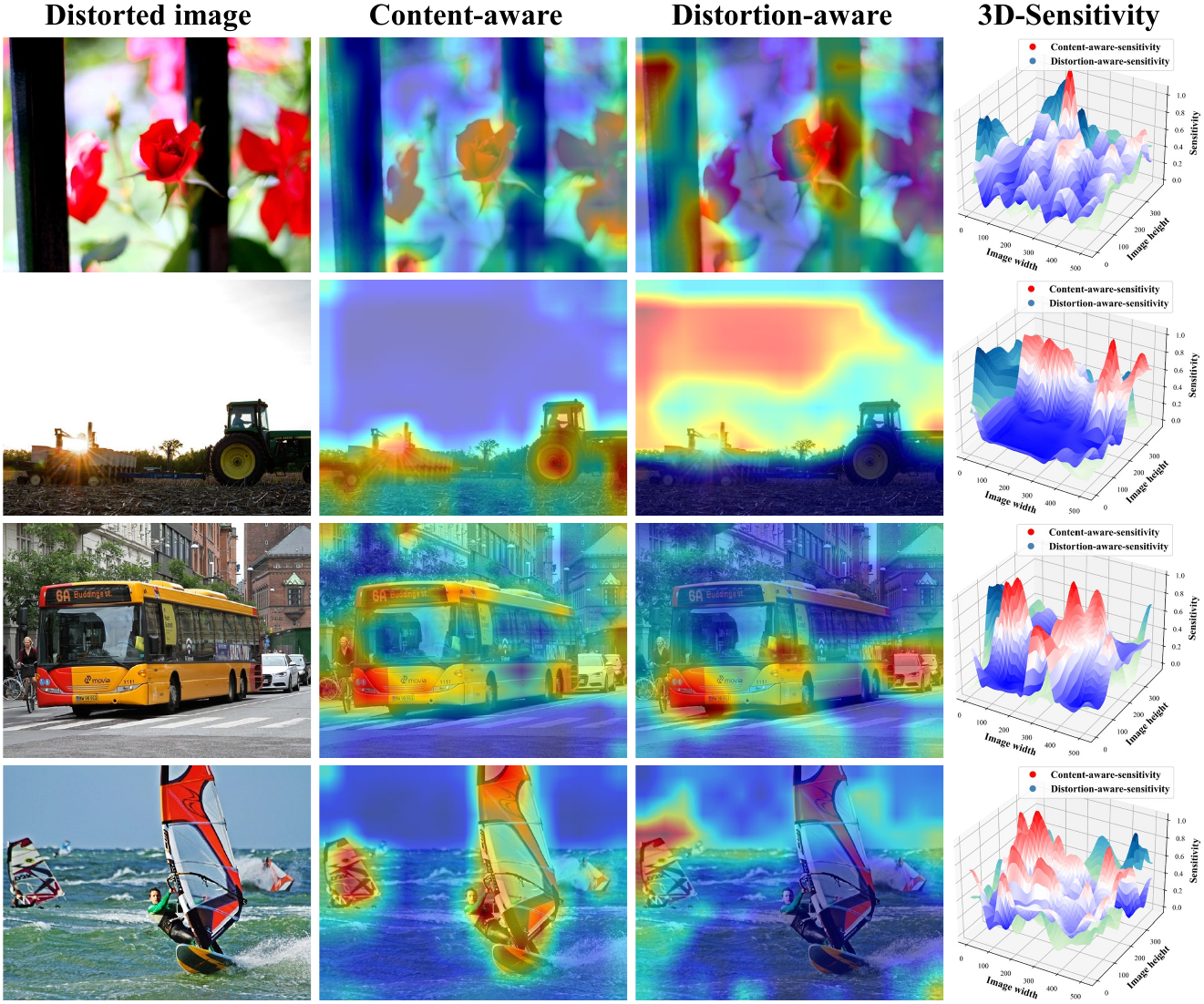}
\caption{Images in the first column: the distorted images in the KonIQ-10K \cite{KonIQ} dataset. Images in the second column: the attention maps from the CAE. Images in the third column: the attention maps from the DAE. Images in the last column: the 3D visualizations derived from columns two and three.}
\label{fig2}
\end{figure}
\begin{figure*}[t]
\centering
\includegraphics[width=0.9\textwidth]{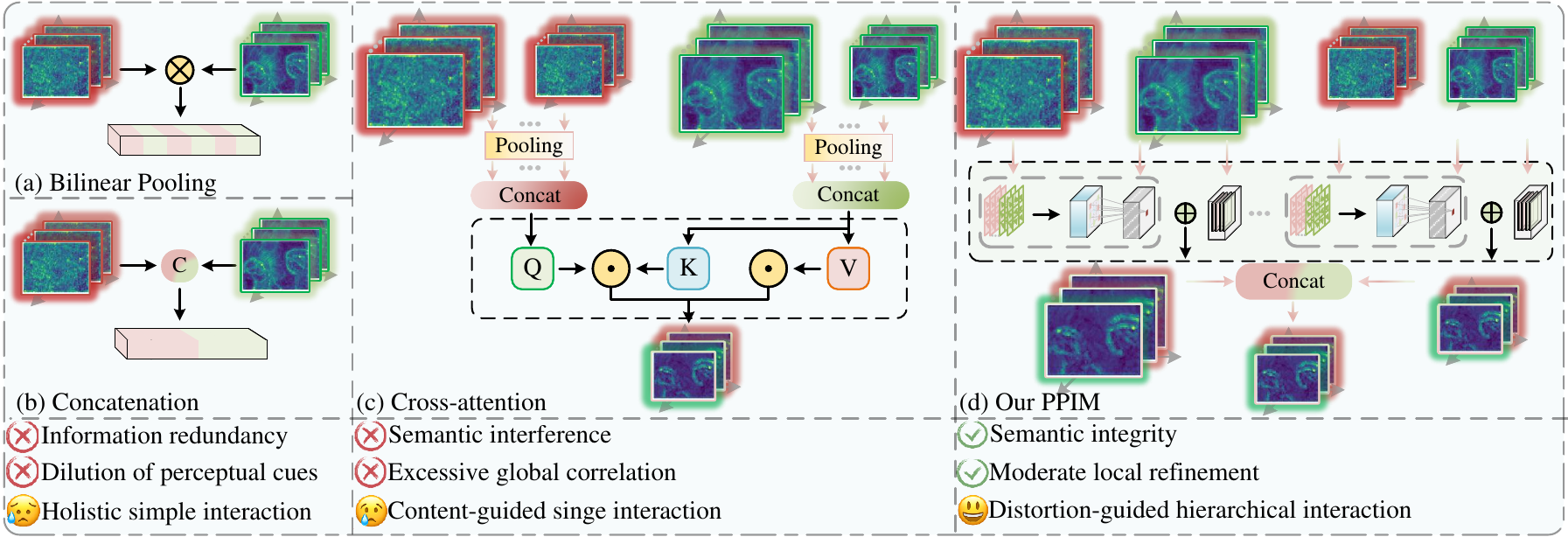}
\caption{Comparison between existing methods and the proposed method for interaction modeling in NR-IQA. Representatives include: (a) DBCNN \cite{zhang2020DBCNN}; (b) Su \textit{et al.} \cite{su2023DisManifold} and Re-IQA \cite{saha2023ReIQA}; (c) CDINet \cite{zheng2024CDINet}; and (d) our PPIM, which is compatible with the interaction properties. More details of (d) can be found in Fig. \ref{fig4}. Feature maps with red glow correspond to distortion features, whereas those with green glow represent to content features.}
\label{fig3}
\end{figure*}

Inspired by the success of deep learning (DL) in various computer vision tasks, many DL-based NR-IQA methods \cite{kang2014CNNIQA, lin2018hallucinated, bosse2018WaDIQaM, wu2020CaHDC, li2021mmmnet, li2021VTIQA, gao2024TCSVT, wang2024TCSVT, shi2025visual} employ an end-to-end strategy to extract image features and predict quality scores. 
Given that existing IQA datasets are insufficient to fully exploit the capabilities of DL models, recent NR-IQA methods primarily follow a pre-training and fine-tuning paradigm. 
Specifically, they utilize convolutional neural networks (CNNs) \cite{li2018, su2020HyperIQA, FLIVE, golestaneh2022TReS} or Transformers \cite{ke2021musiq, qin2023deiqt} pre-trained on large-scale datasets (e.g., ImageNet \cite{deng2009imagenet}) for feature extraction, subsequently fine-tune the backbone and the quality predictor on IQA datasets. 
Unfortunately, these pre-trained models do not perform optimally for IQA because the representations learned from classification task tend to emphasize content information \cite{zhao2023QPT}. 
In contrast to these methods, \cite{su2023DisManifold} and \cite{agnolucci2024ARNIQA} respectively adopt supervised and self-supervised learning to learn the distortion manifold while ignoring image content. 
However, representations learned for IQA should be sensitive to both local distortions and global content, as well as their interactions \cite{zhao2023QPT}. Relying on either aspect alone is insufficient to comprehensively characterize perceptual quality. 
Although some methods attempt to jointly model these two factors, they often fail to capture the complex interactions between them. 
As illustrated in Fig. \ref{fig3}\textcolor{blue}{(a)} and Fig. \ref{fig3}\textcolor{blue}{(b)}, DBCNN \cite{zhang2020DBCNN} fuses content and distortion features through bilinear pooling, whereas Su \textit{et al.} \cite{su2023DisManifold} and Saha \textit{et al.} \cite{saha2023ReIQA} combine them using concatenation. Such simple holistic interaction strategies are prone to introducing redundancy, which in turn dilutes critical perceptual cues. As a result, their prediction accuracy and generalizability are far from ideal.

Unlike prior methods, this work aims to incorporate the interactions between content and distortions into NR-IQA models to better simulate how these factors independently and jointly influence quality perception.
To achieve this, we first select representative distorted images and employ a Content-Aware Encoder (CAE) and a Distortion-Aware Encoder (DAE) to separately extract content-aware and distortion-aware features.
The corresponding attention maps and 3D sensitivity of these features are then visualized to reveal their interaction patterns. 
As shown in the first column of Fig. \ref{fig2}, the central flower in the first image is relatively clear, while the surrounding flowers are noticeably blurred.
In the second image, the background and farming equipment are overexposed, whereas the tractor on the right remains at normal brightness. 
These examples indicate that content and distortions in real-world images are often closely intertwined, with different regions showing varying visual quality. 
In addition, the remaining parts of Fig. \ref{fig2} show that content-aware and distortion-aware features exhibit different spatial sensitivities. 
The former emphasizes structural and semantic information, while the latter highlights areas affected by various degradations. 
This inherent discrepancy poses a challenge for precise interaction modeling. 
Drawing from these observations, we summarize the interaction properties as follows: \textbf{Firstly}, the interactions are highly related to spatial positions, since distortions typically occur in multiple local regions. 
\textbf{Secondly}, the interactions are content-dependent, as human perception of quality can vary with image content \cite{li2018, sun2023stair}.
\textbf{Thirdly}, the feature interaction should be moderate. While distortion-aware features offer valuable degradation cues, excessive interaction may interfere with semantic integrity by disrupting content information. 
\textbf{Finally}, distortions can affect hierarchical features in different ways \cite{wu2020CaHDC}, and visual perception itself follows a hierarchical process. This motivates the incorporation of hierarchical interaction to facilitate a better understanding of quality degradation. 
These key properties form a fundamental basis for interaction modeling, which critically contributes to the development of reliable quality metrics with fine generalizability. 

With these insights, a novel approach, CoDI-IQA (\textbf{Co}ntent-\textbf{D}istortion high-order \textbf{I}nteraction for NR-\textbf{IQA}), is proposed to aggregate local distortion and global content features by exploiting their interactions within a hierarchical interaction framework. In CoDI-IQA, two dedicated encoders are employed to disentangle content-aware and distortion-aware features. Based on the properties we identified, the Progressive Perception Interaction Module (PPIM) is designed to integrate these features through alignment and coarse-to-fine interaction. Specifically, the coarse and fine interaction steps work collaboratively to enhance the interaction representations and facilitate high-order interaction modeling. The former provides basic interaction cues, whereas the latter captures local interaction patterns while preserving semantic integrity. Furthermore, multiple PPIMs are adopted to hierarchically fuse multi-level features to ensure sufficient interaction. To stabilize the interaction process, we also explore a training strategy tailored for CoDI-IQA. Ultimately, the proposed method constructs effective quality-aware representations across diverse distortion scenarios. As shown in Fig. \ref{fig1}, CoDI-IQA achieves significantly improved data efficiency and generalization ability.

Our contributions can be concluded as follows:
\begin{itemize}
\item We analyze the key properties required for interaction modeling and propose a novel NR-IQA method, termed CoDI-IQA. By properly incorporating high-order interaction for quality prediction, the proposed method effectively overcomes the limitations of existing methods in handling interactions between content and distortions.

\item We propose the Progressive Perception Interaction Module (PPIM) to integrate content-aware and distortion-aware features by explicitly modeling their interactions. With compatibility to the desired interaction properties, PPIM combines internal interaction and coarse-to-fine interaction to achieve high-order interaction.

\item To further enhance quality-aware representations, a hierarchical interaction mechanism is introduced to capture interactions at different granularities. Additionally, we explore a specific training strategy to maintain the stability of the interaction process.
    
\item The experimental results on four synthetic IQA datasets and four authentic IQA datasets demonstrate that our method notably outperforms other SOTA competitors. In particular, it shows significant improvements in data efficiency and generalization ability.
\end{itemize}

\section{Related Works}
\subsection{Hand-Crafted-Based NR-IQA}
Early NR-IQA methods \cite{moorthy2011DIIVINE, mittal2012BRISQUE, mittal2012NIQE, saad2012BLIINDS-II, zhang2015ILNIQE, ye2012CORNIA, xu2016HOSA} were primarily designed to handle synthetic distortions. These methods extracted image features using artificially designed feature descriptors and employing simple regression models for quality prediction. Mittal \textit{et al.} \cite{mittal2012BRISQUE} proposed BRISQUE, which used locally normalized luminance coefficients and fit them to Gaussian distributions for feature extraction. NIQE \cite{mittal2012NIQE} extracted features from pristine and distorted images, fit them to a Multivariate Gaussian (MVG) model, and measured image quality by calculating the distance between these models. Zhang \textit{et al.} \cite{zhang2015ILNIQE} developed ILNIQE, which extracted various features from natural image blocks and calculated the overall quality score by averaging the distances between the reference MVG model and the MVG models of distorted blocks. CORNIA \cite{ye2012CORNIA} used K-Means clustering and soft-assignment coding to represent image quality. HOSA \cite{xu2016HOSA} calculated differences in high-order statistics between local features and cluster centers to assess quality. While these methods perform well on synthetically distorted images, they often struggle with the complexity of distortions in natural scenes. This is because manually designed descriptors can only represent a small portion of distortion types and fail to capture content information.

\subsection{Deep Learning-Based NR-IQA}
Recently, advances in deep learning (DL) have evolved NR-IQA from hand-crafted-based to DL-based and achieved significant improvements \cite{kang2014CNNIQA, kim2018DIQA, lin2018hallucinated, ma2017MEON, wu2020CaHDC}. Limited by the sizes of existing IQA datasets, most DL-based NR-IQA methods \cite{kim2017deep, bianco2018, li2018, pan2018, su2020HyperIQA, zhu2020metaiqa, chen2024topiq} used pre-trained CNNs for feature extraction. Li \textit{et al.} \cite{li2018} showed that features obtained from pre-trained ResNet50 could effectively predict quality scores on images in the wild. Su \textit{et al.} \cite{su2020HyperIQA} proposed HyperIQA, which used a pre-trained ResNet50 to extract semantic features, then fed these into a self-adaptive hyper network for evaluation. Zhu \textit{et al.} \cite{zhu2020metaiqa} proposed MetaIQA, a meta-learning-based method that learned a quality prior model and fine-tuned it for unknown distortions. Drawing inspiration from Vision Transformer (ViT) \cite{dosovitskiy2021VIT}, recent developments have integrated Transformers for NR-IQA \cite{ke2021musiq, you2021TRIQ, golestaneh2022TReS, qin2023deiqt}. Ke \textit{et al.} \cite{ke2021musiq} utilized a pre-trained Transformer to extract multi-scale representations from images with the same aspect ratio but different sizes. These methods employed CNNs or Transformers pre-trained on ImageNet \cite{deng2009imagenet}, which tend to extracted features sensitive to global content information. Although Qin \textit{et al.} \cite{qin2023deiqt} attempted to address this by introducing a Transformer decoder, the lack of sensitivity to local distortions still hinders the development of a complete quality perception model.

In contrast, some methods leveraged \cite{madhusudana2022CONTRIQUE, saha2023ReIQA, zhao2023QPT, agnolucci2024ARNIQA} contrastive-based self-supervised learning to pre-train models for NR-IQA. CONTRIQUE \cite{madhusudana2022CONTRIQUE} treated distortion-type classification as the pretext task to obtain distortion representations. ARNIQA \cite{agnolucci2024ARNIQA} modeled the image distortion manifold by maximizing the similarity of image patches with the same degradation but different content. However, relying solely on distortion representations to predict image quality is inconsistent with human perception. To leverage both content and distortion information, Su \textit{et al.} \cite{su2023DisManifold} proposed to learned the distortion manifold and incorporate content information as additional bias. The distortion and semantic embeddings were combined via concatenation. Re-IQA \cite{saha2023ReIQA} used contrastive learning to train two encoders: one for high-level content information and another for low-level quality information. The combined representations from both encoders improved evaluation performance. However, these methods did not fully explore the interactions between content and distortions, which are crucial for understanding their independent and collaborative effects on quality perception. Although CDINet \cite{zheng2024CDINet} employed a content-guided asymmetric cross-attention module (as shown in Fig. \ref{fig3}\textcolor{blue}{(c)}) to capture correlations between features, its excessive global interaction may neglect local interaction patterns and lead to semantic interference. In addition, the quadratic complexity of cross-attention further restricts its capability for hierarchical processing (see Section \ref{secab}\textcolor{blue}{3} for more details). These limitations prevent CDINet from constructing perceptual rules consistent with human visual perception. Our method addresses the above limitations by analyzing the key properties of interaction modeling and heuristically designing the PPIM module (as illustrated in Fig. \ref{fig3}\textcolor{blue}{(d)}) to reveal the underlying impact on image quality caused by the interactions between content and distortions. As a result, our model constructs generalizable and robust quality-aware representations for both synthetic and authentic distortions.
\begin{figure*}[t]
\centering
\includegraphics[width=0.9\textwidth]{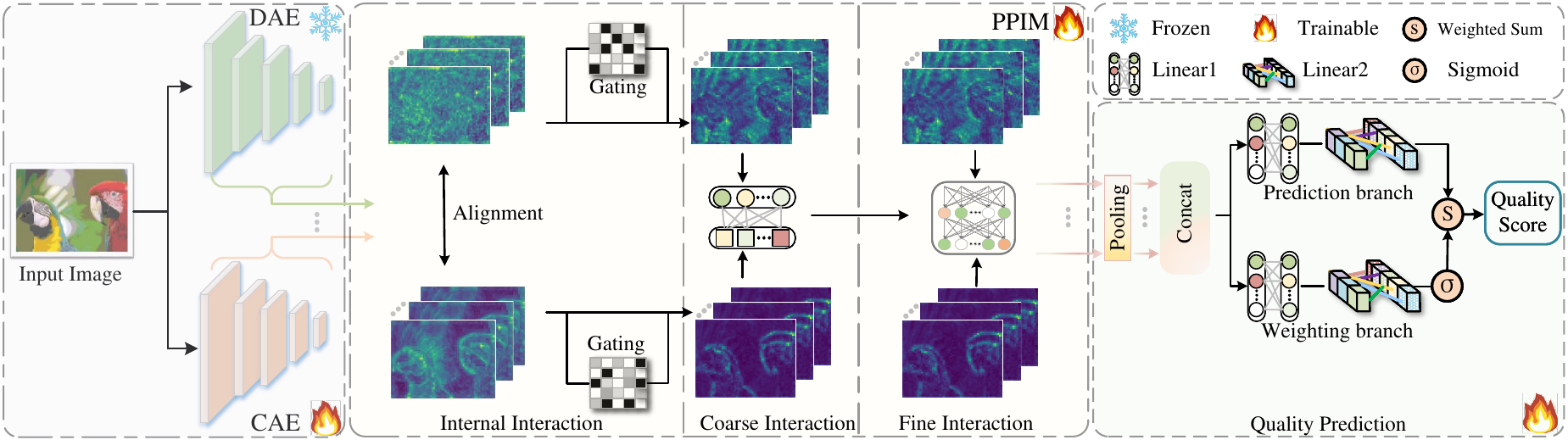}
\caption{The proposed CoDI-IQA involves the CAE and DAE for feature extraction, the PPIM for high-order interaction, and a quality prediction module for generating quality scores.}
\label{fig4}
\end{figure*}
\section{Proposed CoDI-IQA}
The overall architecture of the proposed CoDI-IQA is shown in Fig. \ref{fig4}. It includes three main parts: the feature extraction network, the Progressive Perception Interaction Module (PPIM), and the quality prediction module. Specifically, a Content-Aware Encoder (CAE) and a Distortion-Aware Encoder (DAE) are driven to independently extract content-aware and distortion-aware features from distorted images. Then, the PPIM is designed to integrate these features by exploiting their interactions. To ensure sufficient interaction, multi-level features from both encoders are hierarchically fused by PPIMs at different scales. Finally, a patch-weighted quality prediction module \cite{yang2022MANIQA} is utilized to map the integrated quality-aware representations to quality scores. Furthermore, we explore a tailored training strategy to train the proposed model and maintain interaction stability. The details of each module and the training strategy are introduced as follows.
\begin{table}[t]
\centering
\caption{The Feature Size at Different Stages in the ResNet50, Where “C×H×W” Represents the Channels, Height, and Width of the Feature Size, respectively.}
\label{tab1}
\resizebox{\columnwidth}{!}{
\begin{tabular}{cccc}
\toprule[1.1pt]
Stage & Layer Name & Input Size & Output Size \\
\midrule %
0 & Conv1 & 3 $\times$ H $\times$ W & 64 $\times$ H/2 $\times$ W/2 \\
1 & Conv2\_x & 64 $\times$ H/2 $\times$ W/2 & 256 $\times$ H/4 $\times$ W/4 \\
2 & Conv3\_x & 256 $\times$ H/4 $\times$ W/4 & 512 $\times$ H/8 $\times$ W/8 \\
3 & Conv4\_x & 512 $\times$ H/8 $\times$ W/8 & 1024 $\times$ H/16 $\times$ W/16 \\
4 & Conv5\_x & 1024 $\times$ H/16 $\times$ W/16 & 2048 $\times$ H/32 $\times$ W/32 \\
\bottomrule[1.1pt]
\end{tabular}}
\end{table}
\subsection{Feature Extraction}
\textit{1) Content-Aware Encoder (CAE):} In real-world scenarios, image quality is closely related to its content. Li \textit{et al.} \cite{li2018} pointed out that image-content-aware features can mitigate the impact of content variation on NR-IQA models. These features require heightened sensitivity to image content to enable accurate comprehension of the relationships between content and its underlying semantics. Inspired by this, the CAE is proposed to capture content information. Moreover, to tackle the challenge posed by the vast diversity of image content, the ImageNet \cite{deng2009imagenet} dataset is used to pre-train the CAE to enhance the content-aware ability. ImageNet comprises over 14 million images spanning more than 20,000 distinct categories, most of these images are captured by camera devices and contain abundant authentic distortions. Therefore, directly employing the models pre-trained on ImageNet as the backbone of CAE can simplify the pre-training process. In this work, the CAE is built upon ResNet50 \cite{he2016resnet} or Swin Transformer \cite{liu2021swin}. For clarity, we describe the CAE based on ResNet50 in this section, while the Swin Transformer-based variant is detailed in the supplementary material. Previous methods \cite{su2020HyperIQA, ke2021musiq} have shown the benefits of using multi-scale features extracted from various layers of CNNs for IQA. Motivated by this, we leverage multi-level representations to capture content-aware information at different scales. The feature sizes at different stages of the ResNet50 are listed in Table \ref{tab1}. Multi-scale content-aware features from Stage 0 -- 4 are extracted to facilitate subsequent feature interaction by the PPIM at these stages. This extraction process can be formulated as, 
\begin{equation}
\label{eq1}
[\bm{F}_c^0,\bm{F}_c^1,\bm{F}_c^2,\bm{F}_c^3,\bm{F}_c^4]=\phi_c(I_d),
\end{equation}
where $\phi_c(\cdot)$ indicates the CAE, $I_d\in\mathbb{R}^{3\times H\times W}$ is the input distorted image, and $\bm{F}_c^i\in\mathbb{R}^{C^{i}\times H^{i}\times W^{i}} (i =0, 1, 2, 3, 4)$ indicates the extracted content-aware feature at $i$-th stage.
\begin{figure*}[t]
\centering
\includegraphics[width=0.8\textwidth]{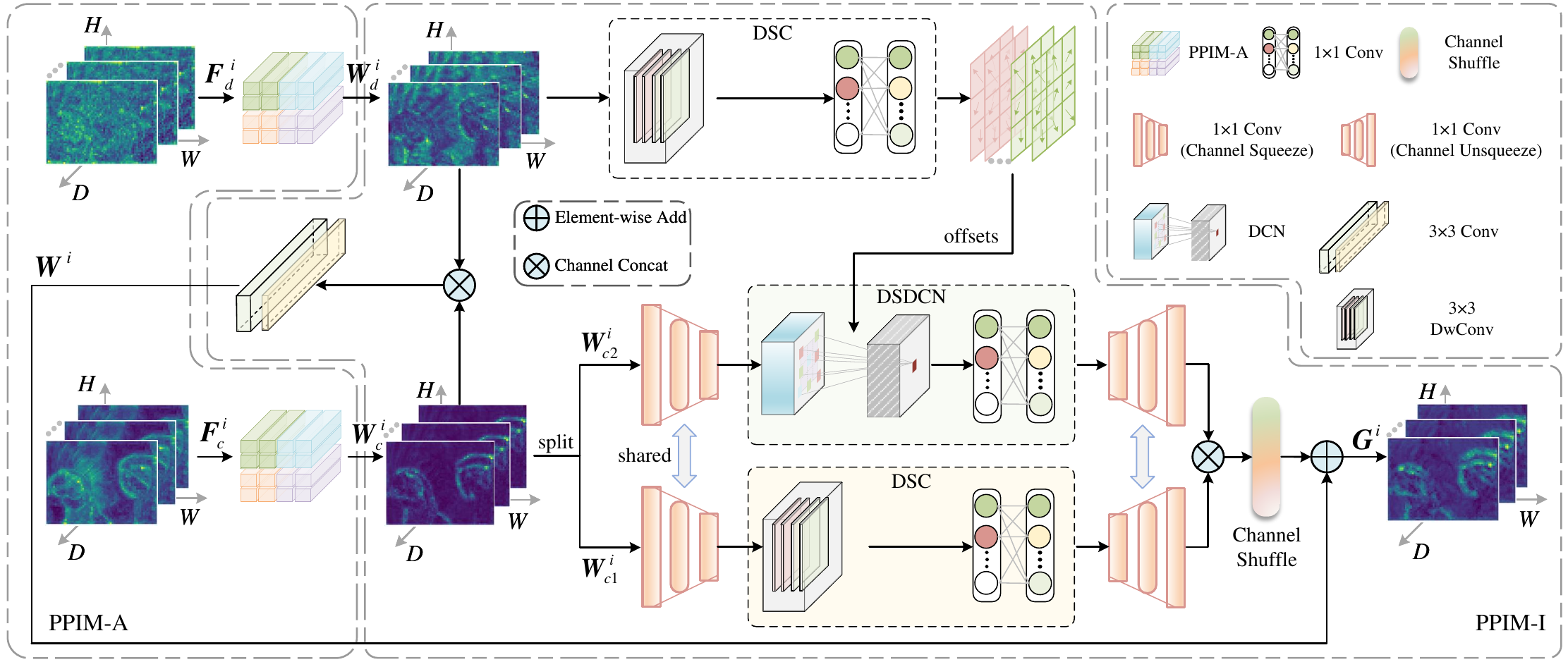}
\caption{The architecture of PPIM. The detailed flowchart outlines the processes involved in high-order interaction.}
\label{fig5}
\end{figure*}

\textit{2) Distortion-Aware Encoder (DAE):} In addition to perceiving image content, it is crucial to capture the degradation patterns in distorted images for constructing a reliable NR-IQA model \cite{su2023DisManifold}. Considering the complexity of distortions in real-world images, training a model exclusively on synthetic images with artificial degradation can only capture limited types and levels of distortion, which are significantly different from authentic conditions. Further, the features extracted from the CAE are sensitive to global content, yet struggle to perceive local distortions. To effectively learn distortion-aware representations, the DAE is built upon ResNet50 pre-trained on the KADIS dataset \cite{KADID} using contrastive loss \cite{agnolucci2024ARNIQA}. By maximizing the similarity of representations between image patches that exhibit the same type of degradation but differ in content, the encoder can recognize various types and degrees of distortion. For this reason, the DAE is capable of capturing degradation patterns to compensate for the limitation of the CAE. Similar to the CAE, multi-scale distortion-aware features from Stage 0 -- 4 are extracted, which is defined as,
\begin{equation}
\label{eq2}
[\bm{F}_d^0,\bm{F}_d^1,\bm{F}_d^2,\bm{F}_d^3,\bm{F}_d^4]=\phi_d(I_d),
\end{equation}
where $\phi_d(\cdot)$ indicates the DAE, $I_d\in\mathbb{R}^{3\times H\times W}$ is the input distorted image, and $\bm{F}_d^i\in\mathbb{R}^{C^{i}\times H^{i}\times W^{i}} (i =0, 1, 2, 3, 4)$ indicates the extracted distortion-aware feature at $i$-th stage.
\subsection{Progressive Perception Interaction Module (PPIM)}
To fully leverage the extracted content-aware and distortion-aware features in a complementary manner, it is essential to consider their interactions when predicting image quality. As outlined in Section \ref{sec:1}, the interactions are content-dependent and closely related to the locations of distortions. It is not straightforward for fusion operations such as addition or concatenation to build the complex interactions needed in scenarios with diverse content and distortions. To address this, the PPIM is proposed to mimic the interactions between content and distortions. By adopting alignment and a coarse-to-fine interaction strategy, the abundant features extracted from both encoders are aligned and fused within the PPIM to obtain meaningful interaction representations. Specifically, the features are first aligned, and a dual-branch structure with a gating mechanism is utilized to achieve internal interaction. Then, the aligned content-aware and distortion-aware features are fused for coarse interaction. Inspired by the adaptive perception process of the human visual system (HVS), a distortion-guided deformable operation is introduced to refine content features, which enables moderate fine interaction within multiple local regions. Finally, the coarse interaction features and the fine interaction features are fused to produce final interaction features in a collaborative manner. Consequently, the PPIM can facilitate high-order interaction to help the model understand how content and distortions independently and collaboratively affect quality perception. Additionally, we apply multiple PPIMs to hierarchically fuse the multi-level features to ensure sufficient interaction. The architecture of the proposed PPIM is illustrated in Fig. \ref{fig5}. For detailed explanation, we divide the PPIM into two parts: feature alignment (PPIM-A) and feature interaction (PPIM-I).
\begin{figure}[t]
\centering
\includegraphics[width=0.7\linewidth]{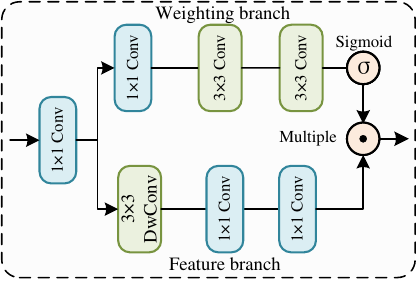}
\caption{The architecture of PPIM-A. The feature branch is an inverted bottleneck block with a consistent dimension $D$. The internal dimension and the output dimension in the weighting branch are set to 64 and 1, respectively. The DW convolution is followed by a BN layer, while other convolutional layers are followed by the GELU activation function.}
\label{fig6}
\end{figure}

\textit{1) Feature Alignment (PPIM-A):} For the features extracted from the $i$-th stage $\bm{F}_c^i$ and $\bm{F}_d^i$ , a $1 \times 1$ convolutional layer is employed to reduce the channels of these features to a unified dimension $D$. This operation not only aligns the features but also decreases the computational complexity for subsequent processes. Considering the discrepancy between these two types of features, a dual-branch structure with different receptive fields is designed to achieve internal interaction. As shown in Fig. \ref{fig6}, a feature branch comprising a $3 \times 3$ depthwise (DW) convolutional layer followed by two $1 \times 1$ convolutional layers is used to enhance the feature representations. Inspired by \cite{yu2019gatedconv}, another weighting branch consisting of one $1 \times 1$ convolutional layer, two $3 \times 3$ convolutional layers, and a sigmoid function is used to compute and apply weight scores to the corresponding features that determines the essential information from both features. Let $\bm{W}_c^i \in\mathbb{R}^{D\times H^{i}\times W^{i}}$ and $\bm{W}_d^i \in\mathbb{R}^{D\times H^{i}\times W^{i}}$ represent the weighted content-aware and distortion-aware features at the $i$-th stage, respectively. These weighted features can be defined as follows,
\begin{equation}
\label{eq3}
\bm{W}_c^i = \sigma(w_3^i(w_1^i(\bm{F}_c^i))) \cdot w_2^i(w_1^i(\bm{F}_c^i)),
\end{equation}
\begin{equation}
\label{eq4}
\bm{W}_d^i = \sigma(w_3^i(w_1^i(\bm{F}_d^i))) \cdot w_2^i(w_1^i(\bm{F}_d^i)),
\end{equation}
where $\sigma$ is the sigmoid function that constrains the weight scores to the range of $[0,1]$, $w_1^i$ indicates the dimension reduction layer, $w_2^i$ represents the feature branch, and $w_3^i$ represents the weighting branch without sigmoid. It is important to note that the PPIM-A for different types of features are independent. A unified description is used here for brevity.

\textit{2) Feature Interaction (PPIM-I):} After acquiring the weighted features $\bm{W}_c^i$ and $\bm{W}_d^i$, a coarse-to-fine interaction strategy is adopted to exploit the complex interactions between them. The coarse and fine interaction steps collaborate to generate precise interaction representations. Specifically, $\bm{W}_c^i$ and $\bm{W}_d^i$ are first fused for coarse interaction through the concatenation operation and a $3 \times 3$ convolutional layer, which can be formulated as,
\begin{equation}
\label{eq5}
\bm{W}^i = w_1^i(\bm{W}_c^i \otimes \bm{W}_d^i),
\end{equation}
where $\bm{W}^i \in\mathbb{R}^{D\times H^{i}\times W^{i}}$ represents the coarse interaction features, $w_1^i$ means the $3 \times 3$ convolutional layer, $\otimes$ indicates the concatenation operation.

Since distortions typically occur in multiple local regions in real-world images, fine interaction needs to account for both distortion locations and content variations. The deformable convolution (DCN) \cite{dai2017deformable} is an ideal tool to fulfill this goal due to its powerful ability to handle deformed spatial locations. Hence, we propose to utilize distortion features to learn offsets and perform deformable operations on content features to focus on distortion regions. To avoid excessive interaction and maintain semantic integrity, $\bm{W}_c^i$ is split into two groups along the channel dimension, $\bm{W}_{c1}^{i}$ and $\bm{W}_{c2}^{i}$. As shown in Fig. \ref{fig5}, The first group utilizes a depthwise separable convolution (DSC), which consists of a $3 \times 3$ depthwise (DW) convolutional layer followed by a $1 \times 1$ convolutional layer, to preserve the content information in $\bm{W}_{c1}^{i}$. Meanwhile, a single depthwise separable and deformable convolution (DSDCN) \cite{qiu2023MBFormer} is employed to adjust $\bm{W}_{c2}^{i}$ to focus on regions impacted by distortions. The DSDCN consists of a $3 \times 3$ DCN followed by a $1 \times 1$ convolutional layer, and another DSC is used to generate the offsets $\Delta p^i$ from $\bm{W}_d^i$. To suppress irrelevant information and reduce computational overhead, a channel squeeze layer is used to down project these features to a smaller dimension $r$.  Adaptive fine interaction is performed in this low-dimensional space, followed by the use of a channel unsqueeze layer to up project the features back to the original dimension. Thereafter, the two groups of features are concatenated and a channel shuffle layer is employed to facilitate inter-group information exchange. Finally, the coarse interaction features $\bm{W}^i$ and fine interaction features are aggregated to achieve coarse-to-fine interaction. The whole process can be formulated as follows,
\begin{equation}
\label{eq6}
\Delta p^i = w_1^i(\bm{W}_d^i),
\end{equation}
\begin{equation}
\label{eq7}
\bm{G}_{1}^{i} = w_3^i(w_4^i(w_2^i(\bm{W}_{c1}^{i}))),
\end{equation}
\begin{equation}
\label{eq8}
\bm{G}_{2}^{i} = w_3^i(w_5^i(w_2^i(\bm{W}_{c2}^{i}), \Delta p^i)),
\end{equation}
\begin{equation}
\label{eq9}
\bm{G}^{i} = \bm{W}^i + w_6^i(\bm{G}_{1}^{i} \otimes \bm{G}_{2}^{i}),
\end{equation}
where $\Delta p^i \in\mathbb{R}^{2N\times H^{i}\times W^{i}}$ represents the learned offsets generated by the DSC $w_1^i$, which are used by the DSDCN $w_5^i$ to adjust the sampling locations, $2N$ denotes the horizontal and vertical offsets for each sampling location. $w_2^i$ and $w_3^i$ indicate the channel squeeze layer and channel unsqueeze layer, respectively, $w_4^i$ indicates the DSC used in first group, $w_6^i$ means the channel shuffle operation, and $\bm{G}^{i} \in\mathbb{R}^{D\times H^{i}\times W^{i}}$ represents the output interaction features.

\begin{table}[t]
\centering
\caption{Summary of Eight Benchmark IQA Datasets.}
\label{tab2}
\resizebox{\columnwidth}{!}{
\begin{tabular}{lcccc}
\toprule[1.1pt]
\multirow{2}{*}{Datasets} & Distorted & Unique & Distortion & \multirow{2}{*}{Label Range} \\
    &  Images &  Contents & Types &  \\
\midrule
LIVE \cite{live} & 779 & 29 & 5 & DMOS [0,100] \\ 
CSIQ \cite{csiq} & 866 & 30 & 6 & DMOS [0,1] \\ 
TID2013 \cite{TID} & 3,000 & 25 & 24 & MOS [0,9] \\
KADID-10K \cite{KADID} & 10,125 & 81 & 25 & MOS [1,5] \\
\midrule
CLIVE \cite{CLIVE} & 1,162 & 1,162 & - & MOS [0,100] \\ 
KonIQ-10K \cite{KonIQ} & 10,073 & 10,073 & - & MOS [1,5] \\
SPAQ \cite{SPAQ}  & 11,125 & 11,125 & - & MOS [0,100] \\
FLIVE \cite{FLIVE}  & 39,810 & 39,810 & - & MOS [0,100] \\
\bottomrule[1.1pt]
\end{tabular}}
\end{table}

\begin{table*}[t]
\centering
\caption{Performance Comparison Measured by Medians of SRCC And PLCC. The Best Result Is Highlighted in \textbf{Bold}, Second-Best Is \underline{Underlined}. Results Maked With $\ast$ Are Obtained From the Retrained Model, And Subsequent Tables Maintain the Same.}
\label{tab3}
\resizebox{\textwidth}{!}{
\begin{tabular}{lcccccccccccccccc}
        \toprule[1.1pt]
        \multirow{2}{*}{\centering Methods\vspace{-5pt}} & \multicolumn{2}{c}{LIVE} & \multicolumn{2}{c}{CSIQ} & \multicolumn{2}{c}{TID2013} & \multicolumn{2}{c}{KADID-10K} & \multicolumn{2}{c}{CLIVE} & \multicolumn{2}{c}{KonIQ-10K} & \multicolumn{2}{c}{SPAQ} & \multicolumn{2}{c}{FLIVE} \\
        \cmidrule(lr){2-3} \cmidrule(lr){4-5} \cmidrule(lr){6-7} \cmidrule(lr){8-9} \cmidrule(lr){10-11} \cmidrule(lr){12-13} \cmidrule(lr){14-15} \cmidrule(lr){16-17}
        & SRCC & PLCC & SRCC & PLCC & SRCC & PLCC & SRCC & PLCC & SRCC & PLCC & SRCC & PLCC & SRCC & PLCC & SRCC & PLCC \\
         \midrule
        BRISQUE \cite{mittal2012BRISQUE} & 0.929 & 0.944 & 0.812 & 0.748 & 0.626 & 0.571 & 0.528 & 0.567 & 0.629 & 0.629 & 0.681 & 0.685 & 0.809 & 0.817 & 0.303 & 0.341 \\
        HOSA \cite{xu2016HOSA} & 0.946 & 0.950 & 0.741 & 0.823 & 0.735 & 0.815 & 0.618 & 0.653 & 0.640 & 0.678 & 0.805 & 0.813 & 0.846 & 0.852 & - & - \\
        WaDIQaM \cite{bosse2018WaDIQaM} & 0.960 & 0.955 & 0.852 & 0.844 & 0.835 & 0.855 & 0.739 & 0.752 & 0.682 & 0.671 & 0.804 & 0.807 & - & - & 0.455 & 0.467 \\
        CaHDC \cite{wu2020CaHDC} & 0.965 & 0.964 & 0.903 & 0.914 & 0.862 & 0.878 & 0.811 & 0.804 & 0.738 & 0.744 & 0.825 & 0.840 & 0.825 & 0.840 & - & - \\
        DBCNN \cite{zhang2020DBCNN} & 0.968 & 0.971 & 0.946 & 0.959 & 0.816 & 0.865 & 0.851 & 0.856 & 0.851 & 0.869 & 0.875 & 0.884 & 0.911 & 0.915 & 0.545 & 0.551  \\
        MetaIQA \cite{zhu2020metaiqa} & 0.960 & 0.959 & 0.899 & 0.908 & 0.856 & 0.868 & 0.762 & 0.775 & 0.835 & 0.802 & 0.887 & 0.856 & - & - & 0.540 & 0.507 \\
        HyperIQA \cite{su2020HyperIQA} & 0.962 & 0.966 & 0.923 & 0.942 & 0.840 & 0.858 & 0.852 & 0.845 & 0.859 & 0.882 & 0.906 & 0.917 & 0.911 & 0.915 & 0.544 & 0.602 \\
        MUSIQ \cite{ke2021musiq} & 0.940 & 0.911 & 0.871 & 0.893 & 0.773 & 0.815 & 0.875 & 0.872 & 0.702 & 0.746 & 0.916 & 0.928 & 0.918 & 0.921 & 0.566 & 0.661 \\
        TReS \cite{golestaneh2022TReS} & 0.969 & 0.968 & 0.922 & 0.942 & 0.863 & 0.883 & 0.859 & 0.858 & 0.846 & 0.877 & 0.915 & 0.928 & - & - & 0.554 & 0.625 \\
        DACNN \cite{pan2022dacnn} & 0.978 & \underline{0.980} & 0.943 & 0.957 & 0.871 & 0.889 & 0.905 & 0.905 & 0.866 & 0.884 & 0.901 & 0.912 & 0.915 & 0.921 & - & - \\
        CONTRIQUE \cite{madhusudana2022CONTRIQUE} & 0.960 & 0.961 & 0.942 & 0.955 & 0.843 & 0.857 & \underline{0.934} & \underline{0.937} & 0.845 & 0.857 & 0.894 & 0.906 & 0.914 & 0.919 & - & - \\
        DEIQT \cite{qin2023deiqt} & \textbf{0.980} & \textbf{0.982} & 0.946 & 0.963 & 0.892 & \underline{0.908} & 0.889 & 0.887 & 0.875 & 0.894 & 0.921 & 0.934 & 0.919 & 0.923 & 0.571 & 0.663 \\
        Su \textit{et al.} \cite{su2023DisManifold} & 0.973 & 0.974 & 0.935 & 0.952 & 0.815 & 0.859 & 0.866 & 0.874 & - & - & - & - & - & - & - & - \\
        Re-IQA \cite{saha2023ReIQA} & 0.970 & 0.971 & 0.947 & 0.960 & 0.804 & 0.861 & 0.872 & 0.885 & 0.840 & 0.854 & 0.914 & 0.923 & 0.918 & 0.925 & - & - \\
        QPT \cite{zhao2023QPT} & - & - & - & - & - & - & - & - & \underline{0.895} & \underline{0.914} & 0.927 & 0.941 & \underline{0.925} & \underline{0.928} & 0.575 & 0.675 \\
        ARNIQA \cite{agnolucci2024ARNIQA} & 0.966 & 0.970 & \textbf{0.962} & \textbf{0.973} & 0.880 & 0.901 & 0.908 & 0.912 & - & - & - & - & 0.905 & 0.910 & - & - \\
        TOPIQ \cite{chen2024topiq} & - & - & - & - & - & - & - & - & 0.870 & 0.884 & 0.926 & 0.939 & 0.921 & 0.924 & \phantom{*}0.574* & \phantom{*}0.657* \\
        CDINet \cite{zheng2024CDINet} & 0.977 & 0.975 & 0.952 & 0.960 & \underline{0.898} & \underline{0.908} & 0.920 & 0.919 & 0.865 & 0.880 & 0.916 & 0.928 & 0.919 & 0.922 & - & - \\
        LoDa \cite{xu2024LoDa} & 0.975 & 0.979 & - & - & 0.869 & 0.901 & 0.931 & 0.936 & 0.876 & 0.899 & \underline{0.932} & 0.944 & \underline{0.925} & \underline{0.928} & \underline{0.578} & \underline{0.679} \\
        \midrule
        \textbf{Ours (R)} & \textbf{0.980} & \underline{0.980} & \underline{0.960} & \underline{0.970} & 0.876 & 0.892 & 0.927 & 0.930 & 0.871 & 0.891 & 0.931 & \underline{0.945} & 0.920 & 0.925 & 0.576 & 0.670 \\
        \textbf{Ours (S)} & 0.978 & \underline{0.980} & 0.957 & 0.967 & \textbf{0.901} & \textbf{0.916} & \textbf{0.936} & \textbf{0.940} & \textbf{0.902} & \textbf{0.917} & \textbf{0.944} & \textbf{0.955} & \textbf{0.927} & \textbf{0.930} & \textbf{0.582} & \textbf{0.685} \\
        \bottomrule[1.1pt]
    \end{tabular}}
\end{table*}

\subsection{Patch-weighted Quality Prediction}
To obtain the final quality-aware feature representation $\bm{G}$, the multi-level interaction features $\bm{G}^{i} (i =0, 1, 2, 3, 4)$ are concatenated. Since the spatial resolutions of these features are inconsistent, the average pooling is first employed to reduce $\bm{G}^{i}$ to the same shape as the highest level features $\bm{G}^{4}$. For brevity, this process is defined as follows,
\begin{equation}
\label{eq10}
\bm{G} = \bm{G}^{0} \otimes \bm{G}^{1} \otimes \bm{G}^{2} \otimes \bm{G}^{3} \otimes \bm{G}^{4},
\end{equation}
where $\bm{G} \in\mathbb{R}^{5D\times H^{4}\times W^{4}}$ is then utilized for quality score generation. We employ a patch-weighted quality prediction module \cite{yang2022MANIQA} instead of a pooling strategy. This ensures consistency between quality prediction and interaction processes, and it accountis for the varying contributions of different image regions to the overall perceived quality. As shown in Fig. \ref{fig4}, this module consists of a prediction and a weighting branch, each implemented using an independent MLP. The prediction branch calculates a quality score for each pixel in the feature map, while the weighting branch computes a weight matrix corresponding to each score. Finally, the overall quality score is obtained through a weighted summation of the individual scores. This process can be expressed as follows,
\begin{equation}
\label{eq11}
Q_{pred}=\frac{\sum s(\bm{G}) * w(\bm{G})}{\sum w(\bm{G})},
\end{equation}
where $s(\bm{G}) \in\mathbb{R}^{H^{4}W^{4} \times 1}$ and $w(\bm{G}) \in\mathbb{R}^{H^{4}W^{4} \times 1}$ denote the outputs of the prediction branch and the weighting branch, respectively, and $*$ means element-wise multiplication. The mean squared error (MSE) loss is utilized to train the proposed method in an end-to-end manner, which is defined as,
\begin{equation}
\label{eq12}
L=\parallel Q_{pred}-Q_{label}\parallel^2,
\end{equation}
where $Q_{pred}$ is the quality score predicted by the proposed model and $Q_{label}$ is the ground-truth quality score derived from subjective experiments.
\subsection{Training Strategy}
To maintain interaction stability, we adopt a specific training strategy for CoDI-IQA. The parameters of the DAE are frozen to ensure that the captured distortion information remains stable throughout training, which is crucial for generating consistent offsets that reflect distortion regions. For the CAE, we employ a commonly used strategy in domain transfer by freezing the batch normalization layers while fine-tuning the remaining parameters, so that the model can adapt to content variations. As shown in Fig. \ref{fig4}, the "trainable" and "frozen" are used to indicate the training status of the modules. Such crafted strategy ensures CoDI-IQA can properly identify distortion locations while its content-adaptive capability, both of which are essential for handling complex interactions.

\section{Experiments}
\subsection{Experimental Setting}
\textit{1) Evaluation Datasets:} We evaluate NR-IQA methods on eight benchmark datasets, including four synthetically distorted datasets: LIVE \cite{live}, CSIQ \cite{csiq}, TID2013 \cite{TID} and KADID-10K \cite{KADID} and four authentically distorted datasets: CLIVE \cite{CLIVE}, KonIQ-10K \cite{KonIQ}, SPAQ \cite{SPAQ}, FLIVE \cite{FLIVE}. The basic information of each dataset are summaried in Table \ref{tab2}.
To ensure consistency, the subjective quality scores of each dataset are scaled to [0,1] using Min-Max normalization.

\textit{2) Evaluation Metrics:} To quantify the performance of each NR-IQA method, two common evaluation metrics are used. Specifically, the Spearman’s rank order correlation coefficient (SRCC) is used to evaluate prediction monotonicity, while the Pearson’s linear correlation coefficient (PLCC) measures prediction accuracy.

\textit{3) Implementation Details:} To train our model, each image is randomly cropped and horizontally flipped into a $384 \times 384$ patch. Importantly, we avoid multiple crops to prevent artificially enlarging the training set. Given the varying image sizes in FLIVE and SPAQ, images are first resized to an appropriate size for training \cite{chen2024topiq}. Specifically, for FLIVE, the shorter side is randomly set between 384 and 416, while for SPAQ, it is set to 448. Regarding the Swin version of CoDI-IQA, images are resized to $384 \times 384$ by default. All datasets are randomly divided into $80\%$ training and $20\%$ testing splits, which are determined based on image content to avoid overlap. To mitigate performance bias, we repeat the training/testing procedure 10 times and report the median results.

We train our model for 200 epochs using the AdamW optimizer with a weight decay of $1 \times 10^{-5}$, and the mini-batch size is set to 8 for all experiments. Early stopping is employed to reduce training time. The initial learning rate is set to $1 \times 10^{-4}$ for synthetic datasets and $3 \times 10^{-5}$ authentic datasets. Following \cite{chen2024topiq}, the cosine annealing scheduler with $T_{max} = 50$ and $\eta_{min} = 0$ is used to adjust the learning rate. The output channel $D$ of the dimension reduction layer is set to 384, and the dimension $r$ after the channel squeeze layer is set to 64. Our model is implemented using PyTorch, and all experiments are performed on an NVIDIA RTX 4090 GPU.
\begin{table*}[t]
\centering
\caption{SRCC and PLCC Results of Individual Distortions on the LIVE Dataset.}
\label{tab4}
\begin{tabular}{lcccccccccc}
\toprule[1.1pt]
                        & \multicolumn{5}{c}{SRCC} & \multicolumn{5}{c}{PLCC} \\
                        \cmidrule(lr){2-6} \cmidrule(lr){7-11}
Methods                  & WN    & GB    & JPEG  & JP2K  & FF    & WN    & GB    & JPEG  & JP2K  & FF    \\
\midrule
BRISQUE \cite{mittal2012BRISQUE} & 0.982 & 0.964 & 0.965 & 0.929 & 0.828 & 0.989 & \textbf{0.965} & 0.971 & 0.940 & 0.894 \\
HOSA \cite{xu2016HOSA} & 0.965 & \textbf{0.972} & 0.921 & 0.920 & 0.934 & 0.959 & \textbf{0.965} & 0.924 & 0.923 & 0.923 \\
WaDIQaM \cite{bosse2018WaDIQaM} & 0.979 & 0.970 & 0.968 & 0.953 & 0.897 & 0.986 & 0.892 & 0.980 & 0.955 & 0.901 \\
CaHDC \cite{wu2020CaHDC}              & 0.978 & 0.951 & 0.970 & 0.948 & 0.898 & 0.982 & 0.955 & 0.953 & 0.973 & 0.913 \\
DBCNN \cite{zhang2020DBCNN}     & 0.980 & 0.935 & 0.972 & 0.955 & 0.930 & 0.988 & 0.956 & 0.986 & 0.967 & 0.961 \\
HyperIQA \cite{su2020HyperIQA}  & 0.982 & 0.926 & 0.961 & 0.949 & 0.934 & 0.982 & 0.921 & 0.962 & 0.946 & 0.916 \\ 
DACNN \cite{pan2022dacnn}  & \textbf{0.986} & 0.959 & 0.974 & 0.962 & 0.949 & \textbf{0.992} & 0.961 & 0.986 & 0.974 & 0.971 \\ 
\midrule
\textbf{Ours (R)}  & 0.981 & 0.960 & \textbf{0.976} & \textbf{0.965} & \textbf{0.965} & \textbf{0.992} & 0.962 & \textbf{0.991} & \textbf{0.982} & \textbf{0.975} \\
\bottomrule[1.1pt]
\end{tabular}
\end{table*}
\begin{table*}[t]
\centering
\caption{SRCC and PLCC Results of Individual Distortions on the CSIQ Dataset.}
\label{tab5}
\begin{tabular}{lcccccccccccc}
\toprule[1.1pt]
& \multicolumn{6}{c}{SRCC} & \multicolumn{6}{c}{PLCC} \\
                        \cmidrule(lr){2-7} \cmidrule(lr){8-13} 
Methods                  & WN    & GB    & JPEG  & JP2K  & PN    & CC    & WN    & GB    & JPEG  & JP2K  & PN    & CC \\
\midrule
BRISQUE \cite{mittal2012BRISQUE}  & 0.723 & 0.820 & 0.806 & 0.840 & 0.378 & 0.824 & 0.742 & 0.891 & 0.828 & 0.878 & 0.496 & 0.835 \\
HOSA \cite{xu2016HOSA} & 0.604 & 0.841 & 0.733 & 0.818 & 0.500 & 0.716 & 0.656 & 0.912 & 0.759 & 0.899 & 0.601 & 0.744 \\
WaDIQaM \cite{bosse2018WaDIQaM} & 0.944 & 0.901 & 0.922 & 0.934 & 0.867 & 0.847 & \textbf{0.956} & 0.916 & 0.934 & 0.957 & 0.886 & 0.873 \\
CaHDC \cite{wu2020CaHDC}              & 0.896 & 0.912 & 0.900 & 0.936 & 0.874 & 0.872 & 0.912 & 0.923 & 0.924 & 0.943 & 0.896 & 0.879 \\
DBCNN \cite{zhang2020DBCNN}    & 0.948 & \textbf{0.947} & 0.940 & 0.953 & 0.940 & 0.870 & \textbf{0.956} & \textbf{0.969} & 0.982 & 0.971 & 0.950 & 0.895 \\
HyperIQA \cite{su2020HyperIQA}  & 0.927 & 0.915 & 0.934 & 0.960 & 0.931 & 0.874 & 0.942 & 0.924 & 0.946 & 0.959 & 0.946 & 0.897 \\ 
DACNN \cite{pan2022dacnn} & \textbf{0.950} & 0.946 & 0.945 & \textbf{0.961} & \textbf{0.956} & 0.885 & 0.908 & 0.95 & 0.982 & 0.960 & 0.946 & 0.921 \\
\midrule
\textbf{Ours (R)} & \textbf{0.950} & 0.943 & \textbf{0.957} & \textbf{0.961} & \textbf{0.956} & \textbf{0.927} & 0.953 & 0.959 & \textbf{0.983} & \textbf{0.972} & \textbf{0.965} & \textbf{0.938} \\
\bottomrule[1.1pt]
\end{tabular}
\end{table*}

\subsection{Performance on Individual Datasets}
\label{seccom}
To demonstrate the superiority of the proposed CoDI-IQA, we compare our method against two hand-crafted-based methods \cite{mittal2012BRISQUE, xu2016HOSA}, five earlier DL-based methods \cite{bosse2018WaDIQaM, wu2020CaHDC, zhang2020DBCNN, zhu2020metaiqa, su2020HyperIQA}, and twelve SOTA methods \cite{ke2021musiq, golestaneh2022TReS, pan2022dacnn, madhusudana2022CONTRIQUE, qin2023deiqt, su2023DisManifold, saha2023ReIQA, zhao2023QPT, agnolucci2024ARNIQA, chen2024topiq, zheng2024CDINet, xu2024LoDa}. The median SRCC and PLCC on eight datasets are presented in Table \ref{tab3}. With the ResNet50 as the CAE, CoDI-IQA achieves highly competitive performance for both synthetic and authentic datasets. In particular, our method notably outperforms CDINet \cite{zheng2024CDINet}, which also aims to model the interactions between content and distortions but employs an asymmetric cross-attention mechanism. However, CoDI-IQA performs worse than CDINet on TID2013, primarily because synthetic distortions are typically globally uniform, such as white Gaussian noise, which affects the entire image and results in a uniform degradation pattern independent of the image content. In such cases, the interplay between content and distortions may be less pronounced locally, thus preventing CoDI-IQA from fully leveraging its advantages. QPT \cite{zhao2023QPT} and LoDa \cite{xu2024LoDa} perform favorably on four authentic datasets. However, QPT demands substantial data and computational resources for pre-training. While LoDa leverages pre-trained ResNet50 and ViT for fine-tuning, it overlooks the fact that classification backbones excessively prioritize content information and remain insensitive to local distortions, let alone their interactions. In contrast, when our method is equipped with a more powerful backbone (Swin Transformer as the CAE), it demonstrates exceptional improvements and achieves the best results in 12 out of 16 comparisons. This indicates that a stronger content-aware capability can help CoDI-IQA better integrate the interactions between content and distortions during the feature fusion process in order to properly simulate their combined impact on image quality. Consistently achieving leading performance is challenging due to the diversity of image content and distortion types across various datasets. These outstanding results highlight the effectiveness and superiority of CoDI-IQA.

To further demonstrate the performance and applicability of CoDI-IQA, we conduct additional experiments on other two widely used IQA benchmark datasets, as well as on datasets from other scenarios, such as night-time images and face images. Details of these experiments can be found in the supplementary material.

\subsection{Performance on Individual Distortions}
To evaluate the performance of CoDI-IQA on different distortion types, we train the model on all distortion types and test it on each individually. LIVE and CSIQ are chosen to conduct the experiments. CoDI-IQA is compared with seven methods \cite{mittal2012BRISQUE, xu2016HOSA, bosse2018WaDIQaM, wu2020CaHDC, zhang2020DBCNN, su2020HyperIQA, pan2022dacnn}. The median SRCC and PLCC for each distortion type in LIVE and CSIQ are reported in Tables \ref{tab4} and \ref{tab5}, respectively. We observe that CoDI-IQA achieves the top performance in 16 out of 22 times, which demonstrates a significant advantage. However, it does not attain the best results on Gaussian blur (GB) and white Gaussian noise (WN). Despite this, CoDI-IQA exhibits more consistent performance across all distortion types. In contrast, other competitors tend to perform inadequately on one or two specific distortion types. This indicates that CoDI-IQA provides greater stability in handling various distortions.

In addition the evaluation on known individual distortions, we also conduct leave-one-distortion-out experiments on the TID2013 and KADID datasets to validate the generalizability of the proposed method to unseen distortions. The results can be found in the supplementary material.
\subsection{Data-Efficient Learning Validation}
Given the substantial costs associated with image annotation, data-efficient learning is crucial for NR-IQA. To investigate this property, we vary the training sample amount from 10\% to 70\% in 10\% intervals while keeping the testing data fixed at 20\% of the total images and completely non-overlapping with the training data. Each experiment is repeated 10 times and the medians of SRCC and PLCC are reported. Three synthetic datasets (LIVE, CSIQ, and KADID-10K) and three authentic datasets (CLIVE, KonIQ-10K, and SPAQ) are chosen to conduct the experiments. The results are detailed in Tables \ref{tab6} and \ref{tab7}. 

On synthetic datasets, CoDI-IQA demonstrates remarkable data efficiency. Specifically, it achieves competitive or even superior performance compared to most competitors in Table \ref{tab3} with only 20\% of the images. When training data exceed 40\%, its performance tends to stabilize and may even slightly decrease. One plausible explanation is that the limited diversity in image content and distortions within synthetic datasets enables CoDI-IQA to effectively model these interactions with fewer samples. Consequently, adding more images with redundant interaction patterns does not substantially improve performance.

From Table \ref{tab7}, we observe that the performance of CoDI-IQA on authentic datasets gradually improves as the amount of training data increases, in contrast to the trend observed in Table \ref{tab6}. This discrepancy arises because real-world images encompass a wider variety of content and distortions, which makes their interactions more complex. Training with additional images allows CoDI-IQA to exploit these interactions to enhance its quality-aware representations. As a result, CoDI-IQA surpasses all methods listed in Table \ref{tab3}, except LoDa, on KonIQ-10K with 60\% images. While CoDI-IQA achieves only comparable results on CLIVE and SPAQ, it is still more efficient than other methods. As shown in the gray rows of Table \ref{tab7}, employing Swin Transformer as the CAE, our method demonstrates admirable data efficiency, as a stronger backbone enables better adaptation to real-world scenarios. Notably, it achieves performance comparable to or better than the top competitors using only 60\%, 30\%, and 60\% images on KonIQ, CLIVE, and SPAQ, respectively, which significantly alleviates the scarcity of training samples for NR-IQA.
\begin{table}[t]
\centering
\caption{Validation With Different Amounts of Training Data on Synthetic Datasets.}
\label{tab6}
\resizebox{\columnwidth}{!}{
\begin{tabular}{ccccccc}
    \toprule[1.1pt]
    \multirow{2}{*}{\centering Amount\vspace{-5pt}} & \multicolumn{2}{c}{LIVE} & \multicolumn{2}{c}{CSIQ} & \multicolumn{2}{c}{KADID-10K} \\
    \cmidrule(lr){2-3} \cmidrule(lr){4-5} \cmidrule(lr){6-7} 
     & SRCC & PLCC & SRCC & PLCC & SRCC & PLCC \\ 
    \midrule
    10\% & 0.968 & 0.967 & 0.944 & 0.954 & 0.917 & 0.919 \\ 
    20\% & 0.974 & 0.974 & 0.946 & 0.957 & 0.921 & 0.922 \\ 
    30\% & 0.975 & 0.976 & 0.947 & 0.956 & 0.923 & 0.926 \\ 
    40\% & 0.976 & 0.976 & 0.954 & 0.960 & \textbf{0.926} & \textbf{0.930} \\ 
    50\% & 0.977 & 0.977 & 0.956 & 0.965 & \textbf{0.926} & 0.929 \\ 
    60\% & 0.977 & 0.977 & 0.951 & 0.963 & \textbf{0.926} & 0.929 \\ 
    70\% & \textbf{0.979} & \textbf{0.979} & \textbf{0.957} & \textbf{0.969} & \textbf{0.926} & 0.929 \\ 
    \bottomrule[1.1pt]
    \end{tabular}}
\end{table}
\begin{table}[t]
\centering
\caption{Validation With Different Amounts of Training Data on Authentic Datasets.}
\label{tab7}
\resizebox{\columnwidth}{!}{
\begin{tabular}{ccccccc}
    \toprule[1.1pt]
    \multirow{2}{*}{\centering Amount\vspace{-5pt}} & \multicolumn{2}{c}{CLIVE} & \multicolumn{2}{c}{KonIQ-10K} & \multicolumn{2}{c}{SPAQ} \\
    \cmidrule(lr){2-3} \cmidrule(lr){4-5} \cmidrule(lr){6-7} 
    & SRCC & PLCC & SRCC & PLCC & SRCC & PLCC \\ 
    \midrule
    10\% & 0.762 & 0.773 & 0.900 & 0.912 & 0.903 & 0.905 \\
    \rowcolor[gray]{0.9}
    &  0.787 & 0.808 & 0.911 & 0.926 & 0.911 & 0.914 \\
    20\% & 0.804 & 0.821 & 0.912 & 0.927 & 0.910 & 0.913 \\
    \rowcolor[gray]{0.9}
    & 0.838 & 0.844 & 0.926 & 0.941 & 0.919 & 0.922 \\
    30\% & 0.826 & 0.842 & 0.918 & 0.932 & 0.912 & 0.916 \\
    \rowcolor[gray]{0.9}
    & 0.854 & 0.864 & 0.933 & 0.943 & 0.920 & 0.923 \\
    40\% & 0.841 & 0.848 & 0.924 & 0.936 & 0.915 & 0.918 \\
    \rowcolor[gray]{0.9}
    & 0.873 & 0.885 & 0.936 & 0.946 & 0.922 & 0.925 \\
    50\% & 0.846 & 0.861 & 0.926 & 0.941 & 0.917 & 0.920 \\
    \rowcolor[gray]{0.9}
    & 0.885 & 0.902 & 0.939 & 0.948 & 0.923 & 0.927 \\
    60\% & 0.853 & 0.865 & 0.928 & 0.942 & 0.918 & 0.922 \\
    \rowcolor[gray]{0.9}
    & 0.891 & 0.912 & 0.941 & 0.952 & 0.925 & 0.928 \\
    70\% & 0.861 & 0.877 & 0.930 & 0.943 & 0.919 & 0.923 \\
    \rowcolor[gray]{0.9}
    & \textbf{0.899} & \textbf{0.915} & \textbf{0.943} & \textbf{0.954} & \textbf{0.926} & \textbf{0.929} \\
    \bottomrule[1.1pt]
    \end{tabular}}
\end{table}

\begin{table*}[t]
\centering
\caption{Cross-Dataset Experiments on Synthetic Datasets.}
\label{tab8}
\begin{tabularx}{0.9\textwidth}{l*{6}{>{\centering\arraybackslash}X}}
    \toprule[1.1pt]
    Training & \multicolumn{3}{c}{LIVE} & \multicolumn{3}{c}{CSIQ}  \\
    \cmidrule(lr){2-4} \cmidrule(lr){5-7} 
    Testing & CSIQ & TID2013 & KADID-10K & LIVE & TID2013 & KADID-10K \\
    \midrule
    DBCNN \cite{zhang2020DBCNN} & 0.758 & 0.524 & 0.481 & 0.877 & 0.540 & 0.463 \\
    MetaIQA \cite{zhu2020metaiqa} & 0.692 & 0.559 & 0.482 & 0.843 & 0.477 & 0.417 \\
    HyperIQA \cite{su2020HyperIQA} & 0.744 & 0.541 & 0.492 & 0.926 & 0.541 & 0.509 \\
    Su \textit{et al.} \cite{su2023DisManifold} & 0.777 & 0.561 & 0.506 & \underline{0.930} & 0.550 & 0.515 \\
    Re-IQA \cite{saha2023ReIQA} & 0.795 & 0.588 & 0.557 & 0.919 & 0.575 & 0.521 \\
    ARNIQA \cite{agnolucci2024ARNIQA} & \textbf{0.904} & \textbf{0.697} & \textbf{0.764} & 0.921 & \textbf{0.721} & \underline{0.735} \\
    \midrule
    \textbf{Ours (R)} & \underline{0.841} & \underline{0.646} & \underline{0.716} & \textbf{0.955} & \underline{0.674} & \textbf{0.752} \\
    \midrule
    Training & \multicolumn{3}{c}{TID2013} & \multicolumn{3}{c}{KADID-10K} \\
    \cmidrule(lr){2-4} \cmidrule(lr){5-7}
    Testing & LIVE & CSIQ & KADID-10K & LIVE & CSIQ & TID2013 \\
    \midrule
    DBCNN \cite{zhang2020DBCNN} & 0.843 & 0.700 & 0.503 & 0.871 & 0.760 & 0.689 \\
    MetaIQA \cite{zhu2020metaiqa} & 0.888 & 0.723 & 0.401 & 0.899 & 0.739 & 0.549 \\
    HyperIQA \cite{su2020HyperIQA} & 0.876 & 0.709 & 0.581 & \underline{0.908} & 0.809 & 0.706 \\
    Su \textit{et al.} \cite{su2023DisManifold} & 0.892 & 0.754 & 0.554 & 0.896 & 0.828 & 0.687 \\
    Re-IQA \cite{saha2023ReIQA} & \underline{0.900} & 0.850 & 0.636 & 0.892 & 0.855 & \underline{0.777} \\
    ARNIQA \cite{agnolucci2024ARNIQA} & 0.869 & \textbf{0.866} & \underline{0.726} & 0.898 & \underline{0.882} & 0.760 \\
    \midrule
    \textbf{Ours (R)} & \textbf{0.941} & \underline{0.852} & \textbf{0.768} & \textbf{0.945} & \textbf{0.913} & \textbf{0.786} \\
\bottomrule[1.1pt]
\end{tabularx}
\end{table*}

\begin{table*}[t]
\centering
\caption{Cross-Dataset Experiments on Authentic Datasets. Here, KonIQ-10K is Referred to KonIQ for Brevity.}
\label{tab9}
\begin{tabularx}{\textwidth}{l*{9}{>{\centering\arraybackslash}X}}
    \toprule[1.1pt]
    Training & \multicolumn{3}{c}{FLIVE} & \multicolumn{2}{c}{KonIQ} & \multicolumn{2}{c}{CLIVE} & \multicolumn{2}{c}{SPAQ} \\
    \cmidrule(lr){2-4} \cmidrule(lr){5-6} \cmidrule(lr){7-8} \cmidrule(lr){9-10}
    Testing & KonIQ & CLIVE & SPAQ & CLIVE & SPAQ & KonIQ & SPAQ & KonIQ & CLIVE \\
    \midrule
    DBCNN \cite{zhang2020DBCNN} & 0.716 & 0.724 & \phantom{*}0.830* & 0.755 & 0.836 & 0.754 & \phantom{*}0.809* & \phantom{*}0.748* & \phantom{*}0.749* \\
    HyperIQA \cite{su2020HyperIQA} & 0.758 & 0.735 & \phantom{*}0.736* & 0.785 & 0.846 & 0.772 & \phantom{*}0.817* & 0.754 & 0.769 \\
    MUSIQ \cite{ke2021musiq} & 0.708 & 0.767 & 0.844 & 0.789 & 0.868 &  \phantom{*}0.583* &  \phantom{*}0.755* & 0.680 & 0.789 \\
    TReS \cite{golestaneh2022TReS} & 0.713 & 0.740 & 0.727 & 0.786 & 0.862 & 0.733 & \phantom{*}0.827* & \phantom{*}0.694* & \phantom{*}0.662* \\
    DEIQT \cite{qin2023deiqt} & 0.733 & 0.781 & - & 0.794 & - & 0.744 & - & - & - \\
    TOPIQ \cite{chen2024topiq} & 0.762 & 0.787 & 0.848 & 0.821 & \underline{0.876} & \phantom{*}0.754* & \phantom{*}0.873* & 0.763 & 0.813 \\
    CDINet \cite{zheng2024CDINet} & - & - & - & 0.750 & - & 0.691 & 0.843 & - & 0.751 \\
    LoDa \cite{xu2024LoDa} & 0.763 & \underline{0.805} & - & 0.811 & - & 0.745 & - & - & - \\
    \midrule
    \textbf{Ours (R)} & \underline{0.806} & 0.792 & \underline{0.858} & \underline{0.825} & 0.874 & \underline{0.804} & \underline{0.877} & \underline{0.799} & \underline{0.824} \\
    \textbf{Ours (S)} & \textbf{0.815} & \textbf{0.816} & \textbf{0.869} & \textbf{0.876} & \textbf{0.890} & \textbf{0.821} & \textbf{0.890} & \textbf{0.817} & \textbf{0.842} \\
\bottomrule[1.1pt]
\end{tabularx}
\end{table*}

\begin{table}[t]
\centering
\caption{Cross-Dataset Experiments on FLIVE.}
\label{tabF}
\resizebox{0.85\columnwidth}{!}{
\begin{tabular}{lccc}
\toprule[1.1pt]
Training  & CLIVE & KonIQ-10K & SPAQ \\
\cmidrule(lr){2-4}
Testing  & FLIVE & FLIVE & FLIVE \\
\midrule
    DBCNN* \cite{zhang2020DBCNN}  & 0.442 & 0.490 & 0.497 \\
    HyperIQA* \cite{su2020HyperIQA} & 0.333 & 0.470 & 0.386 \\
    MUSIQ* \cite{ke2021musiq} & 0.327 & 0.440 & 0.372 \\
    TReS* \cite{golestaneh2022TReS} & 0.331 & 0.437 & 0.362 \\
    TOPIQ* \cite{chen2024topiq} & 0.451 & 0.529 & 0.532 \\
    \midrule
   \textbf{Ours (R)} & \underline{0.484} & \underline{0.541} & \underline{0.547} \\
   \textbf{Ours (S)} & \textbf{0.517} & \textbf{0.573} & \textbf{0.583} \\
    \bottomrule[1.1pt]
\end{tabular}}
\end{table}

\subsection{Generalization Ability Validation}
Cross-dataset evaluation is essential for IQA models as it showcases their ability to generalize across different scenarios. In this section, we evaluate the generalizability of CoDI-IQA by training the model on one dataset and testing it on others without any fine-tuning. We first compare CoDI-IQA with six competitive methods \cite{zhang2020DBCNN, zhu2020metaiqa, su2020HyperIQA, su2023DisManifold, saha2023ReIQA, agnolucci2024ARNIQA} on synthetic datasets. The SRCC results are reported in Table \ref{tab8}. CoDI-IQA obtains the best scores in 7 out of 12 testing items and the second-best in 5 items, which demonstrates superior generalization performance. However, it does not outperform ARNIQA \cite{agnolucci2024ARNIQA} when trained on LIVE. As synthetic images are typically generated from limited pristine images, the content has only a marginal effect on overall image quality. By capturing degradation patterns while disregarding image content, ARNIQA is effective for scenarios with limited content variation. In contrast, CoDI-IQA emphasizes the interplay between content and distortions, which allows it to generalize well on datasets with more diverse content and distortions.

Most NR-IQA methods have undergone limited cross-dataset validation on authentic datasets, which is insufficient to prove their usability in real-world scenarios. We conduct comprehensive cross-dataset validations on four authentic datasets to robustly evaluate the proposed method. Table \ref{tab9} presents the SRCC results of CoDI-IQA in comparison with eight competitors \cite{zhang2020DBCNN, su2020HyperIQA, ke2021musiq, golestaneh2022TReS, qin2023deiqt, chen2024topiq, zheng2024CDINet, xu2024LoDa}. We evaluate FLIVE using its official test split \cite{FLIVE}, which consists of approximately 1.8k images. The comparison results on FLIVE are summarized in Table \ref{tabF}. It can be observed that CoDI-IQA significantly outperforms its competitors across all testing items. Specifically, with ResNet50 as the CAE, CoDI-IQA achieves outstanding generalization performance. When Swin Transformer is used as the CAE, CoDI-IQA shows exceptional improvements. For instance, when trained on KonIQ-10K, it raises the SRCC for CLIVE to 0.876 (+6.2\%). Moreover, the diverse content, sizes, and aspect ratios of FLIVE images make it challenging for other methods to generalize well. In contrast, CoDI-IQA consistently demonstrates stronger generalization ability when tested on FLIVE. These results prove that the proposed method can construct general quality-aware representations that perform well in real-world scenarios.

\begin{figure*}[t]
\centering
\includegraphics[width=0.95\textwidth]{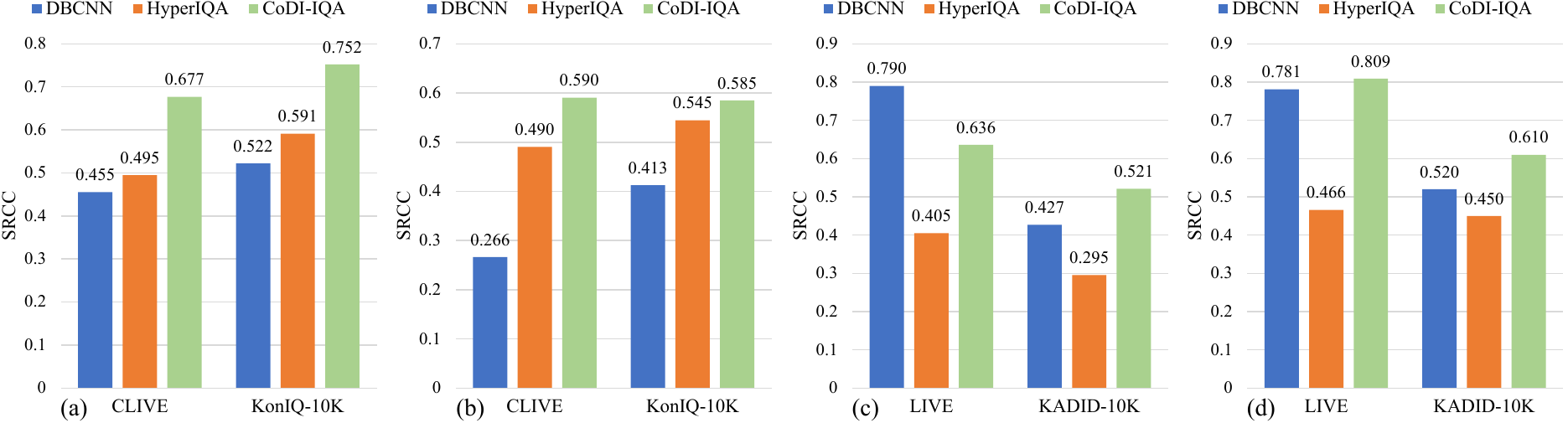}
\caption{Cross-dataset experiments in cross-domain scenarios. (a) and (b) represent models trained on LIVE and KADID-10K, respectively, and tested on CLIVE and KonIQ-10K. (c) and (d) represent models trained on CLIVE and KonIQ-10K, respectively, and tested on LIVE and KADID-10K.}
\label{fig7}
\end{figure*}

\begin{figure*}[t]
\centering
\includegraphics[width=0.95\textwidth]{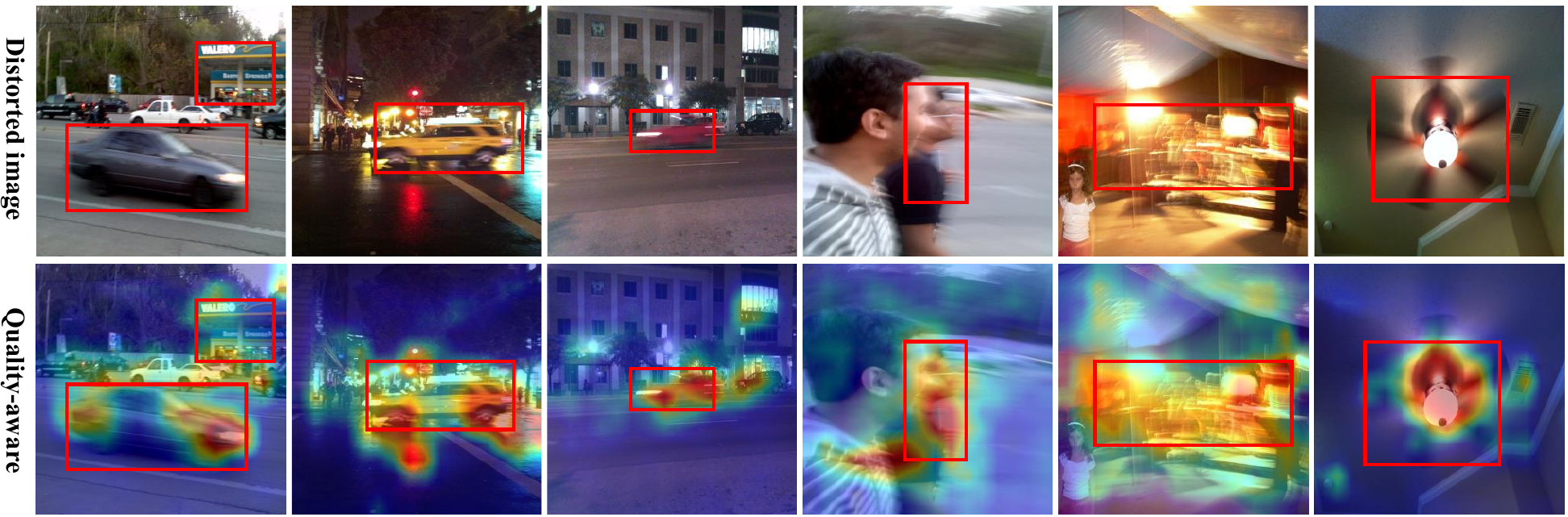}
\caption{Activation maps of CoDI-IQA on the CLIVE dataset show that CoDI-IQA pays more attention to image distortion regions (highlighted in red boxes).}
\label{fig8}
\end{figure*}

Given the large gap between synthetic and synthetic distortions, models trained on synthetic datasets typically struggle to generalize to authentic conditions, and vice versa. To verify the cross-domain generalization ability of CoDI-IQA, we conduct cross-domain experiments with two synthetic datasets (LIVE and KADID-10K) and two authentic datasets (CLIVE and KonIQ-10K). Since most NR-IQA methods do not perform such experiments, we first borrow the SRCC results of DBCNN \cite{zhang2020DBCNN} and HyperIQA \cite{su2020HyperIQA} in synthetic-to-authentic scenarios from \cite{lu2022styleam}. We then retrain DBCNN and HyperIQA to obtain the SRCC scores in authentic-to-synthetic scenarios. The comparison results are shown in Fig. \ref{fig7}. Notably, DBCNN performs better on LIVE, primarily due to its pre-training dataset containing four types of synthetic distortions that also present in LIVE. We are surprised to find that CoDI-IQA consistently exhibits superior generalizability across all cross-domain scenarios, without parameter tuning or domain adaptation. The hypothesized reason is that CoDI-IQA has learned domain-invariant representations by modeling interactions between content and distortions, allowing it to adaptively generalize to different scenarios. We will study this phenomenon in future work. 

To summarize, the above results demonstrate that our method achieves superior generalization ability across different cross-dataset scenarios by considering global content, local distortions, and the interactions between them. This confirms its effectiveness and usability in real-world applications.
\begin{table*}[!ht]
\centering
\caption{Ablation Study Through Cross-Dataset Experiments for Different Components in CoDI-IQA.}
\label{tab10}
\resizebox{0.9\textwidth}{!}{
\begin{tabular}{ccccccccccc}
\toprule[1.1pt]
\multirow{2}{*}{\centering Model index\vspace{-5pt}} & \multirow{2}{*}{\centering CAE\vspace{-5pt}} & \multirow{2}{*}{\centering DAE\vspace{-5pt}} & \multirow{2}{*}{\centering PPIM-A\vspace{-5pt}} & \multirow{2}{*}{\centering PPIM-I\vspace{-5pt}} & \multirow{2}{*}{\centering \shortstack{Trainable \\ Params. (M)}\vspace{-5pt}} &\multicolumn{2}{c}{KADID-10K} & \multicolumn{2}{c}{KonIQ-10K} & \multirow{2}{*}{\centering Average\vspace{-5pt}}\\ 
\cmidrule(lr){7-8} \cmidrule(lr){9-10}
 & & & & & & CSIQ & TID2013 & CLIVE & SPAQ &\\ 
\midrule
\circlednum{1} & \checkmark &  &  &  & 28.65 & 0.783 & 0.664 & 0.767 & 0.853 & 0.767\\
\circlednum{2} &  & \checkmark &  &  & 28.65 & 0.901 & 0.714 & 0.693 & 0.819 & 0.781\\
\circlednum{3} & \checkmark & \checkmark &  &  & 30.15 & 0.887 & 0.747 & 0.808 & 0.860 & 0.826\\
\circlednum{4} & \checkmark & \checkmark & \checkmark &  & 33.77 & 0.896 & 0.751 & 0.812 & 0.862 & 0.830 \\
\circlednum{5} & \checkmark & \checkmark &  & \checkmark & 43.65 & 0.907 & 0.769 & 0.820 & 0.870 & 0.842 \\
\circlednum{6} & \checkmark & \checkmark & \checkmark & \checkmark & 47.27 & \textbf{0.913} & \textbf{0.786} & \textbf{0.825} & \textbf{0.874} & \textbf{0.850} \\
\midrule
\circledlet{a} & \checkmark & \checkmark & \checkmark & Coarse interaction & 47.04 & 0.902 & 0.758 & 0.815 & 0.864 & 0.835\\
\circlednum{b} & \checkmark & \checkmark & \checkmark & Fine interaction & 34.00 & 0.909 & 0.774 & 0.822 & 0.869 & 0.844 \\
\circledlet{c} & \checkmark & \checkmark & \checkmark & Content-guided interaction & 47.27 & 0.906 & 0.765 & 0.817 & 0.867 & 0.839 \\ 
\circlednum{d} & \checkmark & \checkmark & \checkmark & Excessive interaction & 47.37 & 0.908 & 0.779 & 0.815 & 0.865 & 0.842 \\ 
\bottomrule[1.1pt]
\end{tabular}}
\end{table*}
\subsection{Visualization of Attention Map}
To further validate the superiority of CoDI-IQA in capturing quality-related information, we visualize the attention maps of final quality-aware features in Fig. \ref{fig8}, where the ResNet50 version of CoDI-IQA is used. More detailed visualizations are provided in the supplementary material. The results show that CoDI-IQA robustly focuses on the distorted regions while maintaining semantic integrity. We achieve this by facilitating high-order interaction to understand how content and distortions independently and collaboratively affect quality perception, which further helps our model establish quality perception rules consistent with human perception.
\begin{table*}[t]
\centering
\caption{Ablation Study on Hierarchical Feature Interaction. N Represents the Number of PPIMs.}
\label{tab11}
\resizebox{0.75\textwidth}{!}{
\begin{tabular}{cccccccccccc}
\toprule[1.1pt]
\multirow{2}{*}{\centering N\vspace{-5pt}} & \multirow{2}{*}{\centering S4\vspace{-5pt}} & \multirow{2}{*}{\centering S3\vspace{-5pt}} & \multirow{2}{*}{\centering S2\vspace{-5pt}} & \multirow{2}{*}{\centering S1\vspace{-5pt}} & \multirow{2}{*}{\centering S0\vspace{-5pt}} & \multirow{2}{*}{\centering \shortstack{Trainable \\ Params. (M)}\vspace{-5pt}} & \multicolumn{2}{c}{KADID-10K} & \multicolumn{2}{c}{KonIQ-10K} & \multirow{2}{*}{\centering Average\vspace{-5pt}}\\ 
\cmidrule(lr){8-9} \cmidrule(lr){10-11}
& & & & & & & CSIQ & TID2013 & CLIVE & SPAQ &\\ 
\midrule
1 & \checkmark & $\times$ & $\times$ & $\times$ & $\times$ & 28.60 & 0.897 & 0.749 & 0.804 & 0.856 & 0.827 \\ 
2 & \checkmark & \checkmark & $\times$ & $\times$ &  $\times$ & 33.26 & 0.903 & 0.767 & 0.813 & 0.864 & 0.837 \\ 
3 & \checkmark & \checkmark & \checkmark & $\times$ & $\times$ & 37.81 & 0.907 & 0.775 & 0.819 & 0.868 & 0.842 \\ 
4 & \checkmark & \checkmark & \checkmark & \checkmark & $\times$ & 42.47 & 0.910 & 0.780 & 0.824 & 0.872 & 0.847 \\
5 & \checkmark & \checkmark & \checkmark & \checkmark & \checkmark & 47.27 & \textbf{0.913} & \textbf{0.786} & \textbf{0.825} & \textbf{0.874} & \textbf{0.850}
\\
\midrule
1 & \checkmark & \checkmark & \checkmark & \checkmark & \checkmark & 46.48 & 0.906 & 0.784 & 0.818 & 0.869 & 0.844 \\
\bottomrule[1.1pt]
\end{tabular}}
\end{table*}
\subsection{Ablation Study}
\label{secab}
\textit{1) Ablation of the Proposed Components:} The proposed CoDI-IQA consists of three key components: 1) a Content-Aware Encoder (CAE), 2) a Distortion-Aware Encoder (DAE), and 3) a Progressive Perception Interaction Module (PPIM), which is divided into feature alignment (PPIM-A) and feature interaction (PPIM-I). To evaluate the importance of each component, we conduct cross-dataset experiments, where models are trained on KADID-10K and KonIQ-10K, and tested on CSIQ, TID2013, CLIVE, and SPAQ. Cross-dataset tests do not require random splits and lead to a fairer comparison \cite{chen2024topiq}. Therefore, all ablation studies follow the same experimental configuration and adopt the ResNet50 version of CoDI-IQA for consistency. Additionally, all model variants utilize dimension-consistent multi-scale features the same quality prediction module. The SRCC results are reported in Table \ref{tab10}. Results show that using either CAE or DAE alone yields promising performance on synthetic and authentic datasets, respectively. Directly combining CAE and DAE achieves balanced performance on both datasets, but a simple combination may degrade performance. The SRCC increases when all components are present, with PPIM-A slightly improve the performance and PPIM-I bringing the most significant improvement. which proves the effectiveness of PPIM.

\textit{2) Ablation With Different Variants:} To validate the effectiveness of our design, we conduct experiments by constructing four variants of CoDI-IQA: \circledtext{a} removing the fine interaction step in PPIM, \circledlet{b} removing the coarse interaction step in PPIM, \circledtext{c} using content features to generate the offsets for the deformable operation in PPIM, and \circledlet{d} removing the splitting operation in the fine interaction step. From the results presented in Table \ref{tab10}, the following conclusions can be drawn: 1) retaining only the fine interaction step in PPIM can achieve good performance, which highlights the importance of performing content-dependent interaction within multiple distortion locations; 2) the coarse interaction also contributes slightly to the final performance since it further enriches the interaction information through the coarse-to-fine interaction strategy; 3) using content features to generate the offsets in PPIM results in performance degradation, likely because the content-guided refinement does not align well with the properties needed for interaction modeling; 4) the performance degradation caused by excessive interaction is more pronounced on authentic datasets, as semantic interference may lead the model to underestimate quality-aware cues. These findings substantiate the rationality of our design.

\begin{table}[t]
\centering
\caption{Comparison of Different Interaction Methods.}
\label{tabI}
\resizebox{\columnwidth}{!}{
\begin{tabular}{cccccccc}
\toprule[1.1pt]
    \multirow{2}{*}{\centering Methods\vspace{-5pt}} & \multirow{2}{*}{\centering \shortstack{Trainable \\ Params. (M)}\vspace{-5pt}} & \multirow{2}{*}{\centering \shortstack{FLOPs \\ (G)}\vspace{-5pt}} & \multicolumn{2}{c}{KADID-10K} & \multicolumn{2}{c}{KonIQ-10K} & \multirow{2}{*}{\centering Average\vspace{-5pt}}\\
    \cmidrule(lr){4-5} \cmidrule(lr){6-7}
    & & & CSIQ & TID2013 & CLIVE & SPAQ &\\ 
    \midrule
    B-P & 42.50 & 20.45 & 0.901 & 0.767 & 0.817 & 0.864 & 0.837 \\
    ACDI & 43.31 & 10.86 & 0.904  & 0.770 & 0.818 & 0.867 & 0.840 \\
    \textbf{PPIM-I} & 42.47 & 23.90 & \textbf{0.910} & \textbf{0.780} & \textbf{0.824} & \textbf{0.872} & \textbf{0.847} \\
    \bottomrule[1.1pt]
\end{tabular}}
\end{table}

\textit{3) Performance With Different Interaction Methods:} To further demonstrate the effectiveness of our interaction mechanism, we replace PPIM-I with bilinear pooling (B-P) \cite{zhang2020DBCNN} and the asymmetric content-distortion interaction (ACDI) module \cite{zheng2024CDINet} within CoDI-IQA. Specifically, we extract content and distortion features across Stages 1 to 4 while keeping the number of parameters approximately constant to ensure a fair comparison. For ACDI, we reimplement it following the description in \cite{zheng2024CDINet}. As shown in Table \ref{tabI}, bilinear pooling yields the worst performance because it models interactions in a holistic manner, which increases redundancy and subsequently dilutes the perceptual cues within the interaction representations. ACDI shows a slight improvement, as it captures relationships between features from a global perspective. However, its effectiveness in handling non-uniform distortions remains limited. This limitation arises since it lacks adaptive local refinement and struggles to maintain semantic integrity, both of which are crucial for precise interaction modeling. Moreover, hierarchical interaction for ACDI incurs a significant computational cost, which requires 31.97G FLOPs. In contrast, PPIM-I consistently outperforms the competitive methods across all datasets. While our method does not exhibit advantages in FLOPs, it fully accounts for the interaction properties to better cope with complex interactions. These observations provide strong evidence for the capability of the proposed module in proper interaction modeling.
\begin{table}[t]
\centering
\caption{Impact of the Latent Dimensions $D$ And $r$.}
\label{tab12}
\resizebox{\columnwidth}{!}{
\begin{tabular}{cccccccc}
    \toprule[1.1pt]
    \multicolumn{2}{c}{\multirow{2}{*}{\centering Dimensions\vspace{-5pt}}} & \multirow{2}{*}{\centering \shortstack{Trainable \\ Params. (M)}\vspace{-5pt}} &\multicolumn{2}{c}{KADID-10K} & \multicolumn{2}{c}{KonIQ-10K} & \multirow{2}{*}{\centering Average\vspace{-5pt}}\\
    \cmidrule(lr){4-5} \cmidrule(lr){6-7}
    &  & & CSIQ & TID2013 & CLIVE & SPAQ &\\ 
    \midrule
    \multirow{3}{*}{\centering $D$} & 256 & 35.05 & 0.908 & 0.776 & 0.816 & 0.867 & 0.842 \\
    & 384 & 47.27 & \textbf{0.913} & \textbf{0.786} & 0.825 & \textbf{0.874} & \textbf{0.850} \\
    & 512 & 63.92 & 0.912 & 0.785 & \textbf{0.826} & 0.871 & 0.849 \\
    \midrule
    \multirow{3}{*}{\centering $r$} & 48  & 47.22 & 0.910 & 0.780 & 0.819 & 0.870 & 0.845 \\
    & 64  & 47.27 & \textbf{0.913} & \textbf{0.786} & \textbf{0.825} & \textbf{0.874} & \textbf{0.850} \\
    & 80  & 47.33 & 0.911 & 0.784 & 0.823 & 0.868 & 0.847 \\
    \bottomrule[1.1pt]
\end{tabular}}
\end{table}

\textit{4) Performance With Feature Interacted at Different Stages:} In CoDI-IQA, multiple PPIMs are used to hierarchically perform feature interaction between the two encoders across Stage 0 -- 4. To examine the effectiveness of hierarchical interaction, we conduct experiments by gradually incorporating feature interaction from single stage (S4) to five stages (S4 -- S0). The SRCC results are reported in Table \ref{tab11}. Notably, even when interaction occurs only at S4, the performance already surpasses or remains highly competitive with the methods listed in Tables \ref{tab8} and \ref{tab9}. This demonstrates the inherent capability of PPIM for proper interaction modeling, which is our key innovation. Another observation is that as features extracted from S4 to S0 are sequentially interacted, the performance generally increases. Furthermore, when we use a single PPIM to interactively fuse multi-scale content and distortion features, its performance noticeably degrades despite having a comparable number of parameters. This indicates that hierarchical interaction is essential for enhancing the capability of quality-aware representation, rather than merely increasing the number of parameters or aggregating multi-scale features.

\textit{5) Performance With Different Latent Dimensions:} In our Framework, high-level features have twice as many channels as their adjacent low-level features. To align these multi-level features, we first use the dimension reduction layer in PPIM to reduce their channels to unify dimension $D$. During interaction, the channel squeeze layer in PPIM is used to down project these features to a smaller dimension $r$. We conduct ablation studies to analyze the effect of varying $D$ and $r$ on performance. When varying $D$, $r$ is fixed at 64. Conversely, when varying $r$, $D$ is fixed at 384. As shown in Table \ref{tab12}, the latent dimension $D$ significantly affects the number of trainable parameters, whereas $r$ has a minimal impact. However, a larger feature dimension does not necessarily yield better performance. Therefore, we empirically set $D$ to 384 and $r$ to 64 as the default configuration.
\begin{table}[t]
\centering
\caption{Comparison of Different Training Strategies.}
\label{tab13}
\resizebox{\columnwidth}{!}{
\begin{tabular}{ccccccc}
\toprule[1.1pt]
    \multirow{2}{*}{\centering Strategies\vspace{-5pt}} & \multirow{2}{*}{\centering \shortstack{Trainable \\ Params. (M)}\vspace{-5pt}} &\multicolumn{2}{c}{KADID-10K} & \multicolumn{2}{c}{KonIQ-10K} & \multirow{2}{*}{\centering Average\vspace{-5pt}}\\
    \cmidrule(lr){3-4} \cmidrule(lr){5-6}
    & & CSIQ & TID2013 & CLIVE & SPAQ &\\ 
    \midrule
    A & 23.82 & 0.903 & 0.774 & 0.812 & 0.862 & 0.838 \\
    B & 47.27 & \textbf{0.913} & \textbf{0.786} & \textbf{0.825} & \textbf{0.874} & \textbf{0.850} \\
    C & 47.27 & 0.818  & 0.753 & 0.804 & 0.859 & 0.809 \\
    D  & 70.83 & 0.857 & 0.759 & 0.815 & 0.869 & 0.825 \\
    \bottomrule[1.1pt]
\end{tabular}}
\end{table}

\textit{6) Performance With Different Training Strategies:} The training strategy employed for CoDI-IQA is that the DAE is frozen and the other parts are trainable. To evaluate the impact of different training strategies, we compared this strategy with three other variants, which are as follows:
\begin{itemize}
\item A: The CAE and DAE are frozen.
\item B: The CAE is trainable, and the DAE is frozen.
\item C: The CAE is frozen, and the DAE is trainable.
\item D: The CAE and DAE are trainable.
\end{itemize}
According to Table \ref{tab13}, the overall performance ranking is: B \(>\) A \(>\) D \(>\) C, where B is the training strategy used in this work. Strategy B excels because it enables the CAE to adapt to varying image content while preserving the pre-learned distortion information in the DAE, which stabilizes the interaction process and highly compatible with the properties required for PPIM. Strategy C performs the worst as it prevents CoDI-IQA from adequately capturing interactions, despite having the same number of trainable parameters as Strategy B. While Strategy D is slightly less effective on CLIVE and SPAQ, it achieves only moderate performance on CSIQ and TID2013 due to potential instability in the interaction process, which may cause the model to underestimate some quality-related information. Strategy A relies entirely on fixed pre-trained encoders yet demonstrates competitive performance. This suggests that PPIM remains effective in integrating content and distortion features into interaction representations suitable for quality perception, even without the ability to adapt to content variations.
\section{Conclusion}
In this work, a robust NR-IQA method named CoDI-IQA is proposed. By analyzing the interaction properties, we identified the limitations of existing methods in handling the complex interactions between content and distortions. To address this, two dedicated encoders are introduced to disentangle content-aware and distortion-aware features. Subsequently, the Progressive Perception Interaction Module (PPIM) is proposed to facilitate high-order interaction between content and distortions at different granularities in a hierarchical manner. This helps the model reveal the underlying relationship between interaction and perceived quality. Additionally, a crafted training strategy is explored to ensure interaction stability. Experimental results confirm the effectiveness of CoDI-IQA and the interaction modeling capability of PPIM. Thanks to its high data efficiency and strong generalizability, the proposed method is suitable for real-world NR-IQA applications with limited training data and complex distortions. In future work, we plan to further explore cross-scale feature interaction as well as spatiotemporal feature modeling, and extend this framework to video quality assessment.

\bibliographystyle{IEEEtran}  
\bibliography{reference}

\newpage
\twocolumn[
\begin{center}
    {\LARGE Supplementary Material}
\end{center}
\vspace{1cm}]

\section{Introduction}
This document serves as the supplementary material for our manuscript, \textit{Content-Distortion High-Order Interaction for Blind Image Quality Assessment}. Please note that all bibliography indexes referenced here correspond to those in the main manuscript.

\begin{table}[h]
\centering
\caption{The Feature Size at Different Stages in the Swin-Base Transformer, Where “C×H×W” Represents the Channels, Height, and Width of the Feature Size, respectively.}
\label{tabs1}
\resizebox{\columnwidth}{!}{
\begin{tabular}{cccc}
\toprule[1.1pt]
Stage & Layer Name & Input Size & Output Size \\
\midrule %
1 & Layer1 & 3 $\times$ H $\times$ W & 128 $\times$ H/4 $\times$ W/4 \\
2 & Layer2 & 128 $\times$ H/4 $\times$ W/4 & 256 $\times$ H/8 $\times$ W/8 \\
3 & Layer3 & 256 $\times$ H/8 $\times$ W/8 & 512 $\times$ H/16 $\times$ W/16 \\
4 & Layer4 & 512 $\times$ H/16 $\times$ W/16 & 1024 $\times$ H/32 $\times$ W/32 \\
\bottomrule[1.1pt]
\end{tabular}}
\end{table}

\section{More Description of CoDI-IQA}
In the main manuscript, we state that the CAE is built upon ResNet50 \cite{he2016resnet} or Swin Transformer \cite{liu2021swin}, and we provide a detailed description of the ResNet50 version of CoDI-IQA. For the Swin-based CoDI-IQA, the Swin-Base Transformer is adopted as the backbone for the CAE, while the DAE remains unchanged. Specifically, the Swin-Base Transformer is pre-trained on ImageNet-22k and fine-tuned on ImageNet-1k. The feature sizes at different stages of it are summarized in Table \ref{tabs1}. Multi-scale content-aware features from Stage 1 -- 4 are extracted, which can be formulated as, 
\begin{equation}
\label{eqs1}
[\bm{F}_c^1,\bm{F}_c^2,\bm{F}_c^3,\bm{F}_c^4]=\phi_c(I_d),
\end{equation}
where $\bm{F}_c^i\in\mathbb{R}^{C^{i}\times H^{i}\times W^{i}} (i = 1, 2, 3, 4)$ indicates the extracted content-aware feature.

Correspondingly, Multi-scale distortion-aware features are also extracted from the DAE, as outlined below:
\begin{equation}
\label{eqs2}
[\bm{F}_d^1,\bm{F}_d^2,\bm{F}_d^3,\bm{F}_d^4]=\phi_d(I_d),
\end{equation}
where $\bm{F}_d^i\in\mathbb{R}^{C^{i}\times H^{i}\times W^{i}} (i = 1, 2, 3, 4)$ indicates the extracted distortion-aware feature.

After obtaining these multi-scale features, four PPIMs are employed to hierarchically integrate them to generate the interaction features. The interaction process is described in detail in the main manuscript and is not elaborated here. The multi-level interaction features $\bm{G}^{i} (i = 1, 2, 3, 4)$ are concatenated as,
\begin{equation}
\label{eqs3}
\bm{G} = \bm{G}^{1} \otimes \bm{G}^{2} \otimes \bm{G}^{3} \otimes \bm{G}^{4},
\end{equation}
where $\bm{G} \in\mathbb{R}^{4D\times H^{4}\times W^{4}}$ is then utilized for quality score generation, with $D$ consistently set to 384.

Compared to the ResNet50 version of CoDI-IQA, although the features from stage 0 are omitted, the more powerful classification backbone enables the proposed PPIM to better capture and leverage the interplay between content and distortions. As a result, the Swin-based CoDI-IQA ultimately shows notable improved performance. This  demonstrates the flexibility of the proposed interaction framework, which can adapt to heterogeneous encoder combinations.

\begin{table}[h]
\centering
\caption{Performance Comparison Measured SRCC And PLCC. The Best Result Is Highlighted in \textbf{Bold}, Second-Best Is \underline{Underlined}. Results Maked With $\ast$ Are Obtained From the Retrained Model, And Subsequent Tables Maintain the Same.}
\label{tabs2}
\resizebox{\columnwidth}{!}{
\begin{tabular}{lcccccc}
    \toprule[1.1pt]
    \multirow{2}{*}{\centering Methods\vspace{-5pt}} & \multicolumn{2}{c}{BID} & \multicolumn{2}{c}{CID2013} & \multicolumn{2}{c}{FLIVE} \\
    \cmidrule(lr){2-3} \cmidrule(lr){4-5} \cmidrule(lr){6-7} 
     & SRCC & PLCC & SRCC & PLCC & SRCC & PLCC \\ 
    \midrule
    BRISQUE \cite{mittal2012BRISQUE} & 0.562 & 0.593 & 0.629 & 0.642 & 0.288 & 0.373 \\ 
    HOSA \cite{xu2016HOSA} & 0.721 & 0.736 & 0.679 & 0.683 & - & - \\ 
    WaDIQaM \cite{bosse2018WaDIQaM} & 0.653 & 0.636 & 0.696 & 0.712 & 0.434 & 0.430 \\ 
    DBCNN \cite{zhang2020DBCNN} & 0.845 & 0.859 & 0.828 & 0.839 & 0.554 & 0.652 \\ 
    HyperIQA \cite{su2020HyperIQA} & 0.869 & 0.878 & 0.871 & 0.885 & 0.535 & 0.623 \\ 
    MUSIQ \cite{ke2021musiq} &   -  &   -   &  -   &   -   & \textcolor{gray}{0.646} & \textcolor{gray}{0.739} \\
    CONTRIQUE \cite{madhusudana2022CONTRIQUE} &   -  &   -   &  -   &   -   & 0.580 & 0.641 \\
    Re-IQA \cite{saha2023ReIQA} &   -  &   -   &  -   &   -   & \underline{0.645} & \underline{0.733} \\
    QPT \cite{zhao2023QPT} & 0.888 & \textbf{0.911} &  -   &   -   & 0.610 & 0.677 \\
    ARNIQA \cite{agnolucci2024ARNIQA} &   -  &   -   &  -   &   -   & 0.595 & 0.671 \\
    TOPIQ* \cite{chen2024topiq} & 0.882 & 0.897 & \underline{0.954} & \underline{0.955} & 0.633 & 0.709 \\
    CDINet \cite{zheng2024CDINet} & 0.874 & \underline{0.899} &   -  &   -   &  -   &   -       \\
    LoDa \cite{xu2024LoDa} & 0.885 & 0.883 &   -  &   -   &  -   &   -       \\
    \midrule
    \textbf{Ours (R)}    & \underline{0.892} & 0.894 & \textbf{0.955} & \textbf{0.956} & 0.636 & 0.708 \\
    \textbf{Ours (S)}    & \textbf{0.894} & \textbf{0.911} & 0.952 & 0.950 & \textbf{0.650} & \textbf{0.734} \\
    \bottomrule[1.1pt]
    \end{tabular}}
\end{table}

\section{More Experiments}
\subsection{Performance on More IQA Datasets}
Beyond the eight standard IQA benchmark datasets discussed in the main manuscript, certain NR-IQA methods have also performed experiments on the BID \cite{ciancio2011BID} and CID2013 \cite{virtanen2014cid2013} datasets. BID includes 586 images with various authentic blur distortions, such as simple motion blur, complex motion blur, and out-of-focus blur. CID2013 is an authentic IQA that comprises 480 images taken from eight distinct scenes under laboratory conditions.  Additionally, some approaches evaluate their performance on FLIVE using its official train/test split \cite{FLIVE}, which is not included in the main manuscript. To further validate the effectiveness of the proposed method, we conduct additional experiments on BID, CID2013, and FLIVE datasets. The compared results are presented are presented in Table \ref{tabs2}. Note that MUSIQ utilizes additional 90K training patches to boost its performance on FLIVE, whereas all other methods are trained solely on 30K images. It can be observed that CoDI-IQA consistently achieves superior performance when evaluated on authentic blur distortions as well as on diverse real-world images. These results indicate its robustness across various scenarios.
\begin{table}[h]
\centering
\caption{More Cross-Dataset Experiments on Authentic Datasets. \\ Here, KonIQ-10K is Referred to KonIQ for Brevity.}
\label{tabs3}
\resizebox{\columnwidth}{!}{
\begin{tabular}{lcccccc}
    \toprule[1.1pt]
     Training & KonIQ & SPAQ & CLIVE & \multicolumn{2}{c}{BID} & \multirow{2}{*}{\centering Average\vspace{-5pt}} \\
    \cmidrule(lr){2-6} 
     Testing & BID & BID & BID & CLIVE & KonIQ & \\ 
    \midrule
    DBCNN \cite{zhang2020DBCNN}  & 0.816 & - & 0.762 & 0.725 & 0.724 & 0.757\\ 
    HyperIQA \cite{su2020HyperIQA} & 0.819 & - & 0.756 & 0.770 & 0.688 & 0.758 \\ 
    QPT \cite{zhao2023QPT} & 0.825 & - & 0.845 &   -   & - & - \\
    TOPIQ* \cite{chen2024topiq} & 0.847 & 0.798 & 0.895 & 0.802 & 0.662 & 0.801  \\
    CDINet \cite{zheng2024CDINet} & 0.840 & 0.771 & 0.862 & 0.693 & 0.694 & 0.772 \\
    LoDa \cite{xu2024LoDa} & \underline{0.850} & -  & 0.890 & 0.805 & 0.733 & 0.820\\
    \midrule
    \textbf{Ours (R)}    & \underline{0.850} & \underline{0.817} & \underline{0.902} & \underline{0.828} & \underline{0.757} & \underline{0.831} \\
    \textbf{Ours (S)}    & \textbf{0.874} & \textbf{0.847} & \textbf{0.903} & \textbf{0.853} & \textbf{0.799} & \textbf{0.855} \\
    \bottomrule[1.1pt]
    \end{tabular}}
\end{table}
\subsection{Generalization Ability Validation}
Additional cross-dataset experiments are conducted to provide a more comprehensive comparison of CoDI-IQA against existing methods. BID and three authentic datasets (CLIVE \cite{CLIVE}, KonIQ-10K \cite{KonIQ}, and SPAQ \cite{SPAQ}) mentioned in the main manuscript are chosen for evaluation. The SRCC results are shown in Table \ref{tabs3}. Obviously, the proposed method achieve the best results across all test items. In particular, CoDI-IQA significantly outperforms CDINet \cite{zheng2024CDINet}, with substantial increases in average SRCC of 7.6\% and 10.8\%, respectively. This reaffirms the effectiveness of our approach in modeling interactions. LoDa \cite{xu2024LoDa} performs second only to CoDI-IQA, as it injects local distortion features from ResNet50 into ViT, which can be interpreted as an implicit interaction modeling between local distortions and global content. However, since both ResNet50 and ViT used in LoDa are pre-trained on image classification tasks, the extracted features tend to overly focus on content information while remaining insensitive to distortion information, which in turn hinders LoDa to effectively to reflect the underlying interaction patterns. In contrast, CoDI-IQA incorporates a DAE that is specifically pre-trained to learn the image distortion manifold. The DAE collaborates with the CAE to disentangle content and distortion features, and this explicit separation provides a stronger foundation for subsequent feature interaction.
\begin{table*}[b]
\centering
\caption{SRCC And PLCC Results of Further Training on Target Datasets.}
\label{tabs4}
\resizebox{0.8\textwidth}{!}{
\begin{tabular}{cccccccccc}
\toprule[1.1pt]
\multirow{2}{*}{\centering Methods\vspace{-5pt}} & \multirow{2}{*}{\centering Pre-trained\vspace{-5pt}} & \multicolumn{4}{c}{BID} & \multicolumn{4}{c}{CLIVE} \\
\cmidrule(lr){3-6} \cmidrule(lr){7-10}
& & SRCC & $\Delta$ & PLCC & $\Delta$ & SRCC & $\Delta$ & PLCC & $\Delta$ \\
\midrule
\multirow{5}{*}{Ours (R)} 
& BID   & 0.892 & -- & 0.894 & -- & 0.881 & +1.15\% & 0.894 & +0.34\% \\
& CLIVE & \textbf{0.924} & \textcolor{blue}{\textbf{+3.58\%}} & \textbf{0.938} & \textcolor{blue}{\textbf{+4.92\%}} & 0.871 & -- & 0.891 & -- \\
& KonIQ & 0.900 & +0.90\% & 0.911 & +1.90\% & \textbf{0.891} & \textcolor{blue}{\textbf{+2.30\%}} & \textbf{0.907} & \textcolor{blue}{\textbf{+1.80\%}} \\
& SPAQ  & 0.899 & +0.78\% & 0.911 & +1.90\% & 0.882 & +1.26\% & 0.904 & +1.46\% \\
& FLIVE & 0.895 & +0.34\% & 0.901 & +0.78\% & 0.874 & +0.34\% & 0.889 & \textcolor{gray}{\textbf{-0.22\%}} \\
\midrule
\multirow{5}{*}{Ours (S)} 
& BID   & 0.894 & -- & 0.911 & -- & 0.911 & +1.00\% & 0.929 & +1.31\% \\
& CLIVE & \textbf{0.926} & \textcolor{blue}{\textbf{+3.59\%}} & \textbf{0.943} & \textcolor{blue}{\textbf{+3.51\%}} & 0.902 & -- & 0.917 & -- \\
& KonIQ & 0.915 & +2.35\% & 0.928 & +1.87\% & \textbf{0.921} & \textcolor{blue}{\textbf{+2.11\%}} & \textbf{0.934} & \textcolor{blue}{\textbf{+1.85\%}} \\
& SPAQ  & 0.912 & +2.01\% & 0.927 & +1.76\% & 0.908 & +0.67\% & 0.924 & +0.76\% \\
& FLIVE & 0.896 & +0.22\% & 0.915 & +0.44\% & 0.899 & \textcolor{gray}{\textbf{-0.33\%}} & 0.914 & \textcolor{gray}{\textbf{-0.33\%}} \\
\bottomrule[1.1pt]
\end{tabular}}
\end{table*}
\subsection{Pre-trained and Fine-tuned on Target Datasets}
As outlined in the main manuscript, the proposed method demonstrates a remarkable data-efficient learning capability, which significantly alleviates the challenge posed by the scarcity of training samples in NR-IQA. However, the limitation in the number of labeled images still persists. For example, BID contains only 586 images, which is approximately one-seventeenth the size of the KonIQ-10K. For real-world scenarios, such a limited size may hinder the model from comprehensively learning the impact of diverse distortions and content variations on image quality. Fortunately, CoDI-IQA exhibits strong generalization ability on these datasets. As shown in Tables \ref{tabs2} and \ref{tabs3}, when trained on CLIVE, the cross-dataset performance on BID already surpasses the performance achieved by training directly on BID. This observation raises an important question of whether such well-generalized models can be fine-tuned on the target dataset to further improve their performance. To this end, we perform experiments by fine-tuning the pre-trained CoDI-IQA models on BID and CLIVE. The SRCC and PLCC results are listed in Table \ref{tabs4}. 

As we can see, fine-tuning CoDI-IQA on BID after pre-training on four larger datasets leads to varying degrees of performance improvement, with the model pre-trained on CLIVE achieving the most significant gain. A similar pattern is observed when fine-tuning on CLIVE, where the model pre-trained on KonIQ-10K achieves the highest improvement, while using FLIVE for pre-training results in slighty performance degradation. Moreover, both versions of CoDI-IQA perform favorably when pre-trained on CLIVE and fine-tuned on BID, and vice versa. From these phenomenon, two key conclusions can be draw. Firstly, a good pre-trained model is crucial for improving performance on small datasets. CoDI-IQA benefits from its ability to capture the complex interactions between content and distortions, and it reveals how these interactions influence perceptual quality. This understanding leads to more robust quality-aware initializations that support effective knowledge transfer. Secondly, larger datasets do not always guarantee better transfer results. Although FLIVE is the largest dataset, the wide range of images it contains results in a substantial domain gap between FLIVE and other datasets, which may negatively affect transferability.

\begin{table*}[t]
\centering
\caption{Leave-One-Distortion-Out Performance Comparison on the TID2013 Dataset.}
\label{tabs5}
\resizebox{\textwidth}{!}{
\begin{tabular}{l|cccccccccc|c}
\toprule[1.1pt]
Dist. Type & BRISQUE & HOSA & WaDIQaM & DBCNN & MetalQA & HyperIQA & MUSIQ* & TReS* & Su \textit{et al.} & ARNIQA* & \textbf{Ours (R)}\\ 
\midrule
AGN   & 0.9356 & 0.7582 & 0.9080 & 0.9680 & 0.9473 & 0.9590 & 0.9345 & 0.9670 & \textbf{0.9698} & 0.9500 & 0.9689 \\
ANC   & 0.8114 & 0.4670 & 0.8700 & 0.9231 & 0.9240 & 0.9201 & 0.8915 & \textbf{0.9441} & 0.9247 & 0.9256 & 0.9338\\
SCN   & 0.5457 & 0.6246 & 0.8802 & 0.9704 & 0.9534 & 0.9693 & 0.9406 & 0.9736 & 0.9708 & 0.9368 & \textbf{0.9739}\\
MN    & 0.5852 & 0.5125 & 0.8065 & 0.8253 & 0.7277 & 0.7606 & 0.6875 & 0.7096 & 0.8438 & 0.8248 & \textbf{0.8454}\\
HFN   & 0.8965 & 0.8285 & 0.9314 & 0.9520 & 0.9518 & 0.9597 & 0.9330 & 0.9650 & 0.9611 & 0.9468 & \textbf{0.9696}\\
IN    & 0.6559 & 0.1889 & 0.8779 & 0.7256 & 0.8653 & 0.7730 & 0.8437 & 0.7868 & 0.6849 & 0.8446 & \textbf{0.9228} \\
QN    & 0.6555 & 0.4145 & 0.8541 & 0.8807 & 0.7454 & 0.8622 & 0.7218 & 0.8706 & 0.8074 & 0.9027 & \textbf{0.9136}\\
GB    & 0.8656 & 0.7823 & 0.7520 & 0.9619 & 0.9767 & 0.9704 & 0.9087 & 0.9749 & 0.9775 & 0.9293 & \textbf{0.9799}\\
DEN   & 0.6143 & 0.5436 & 0.7680 & 0.9406 & 0.9383 & 0.9604 & 0.8512 & 0.9521 & 0.9409 & 0.8936 & \textbf{0.9642}\\
JPEG  & 0.5186 & 0.8318 & 0.7841 & 0.9434 & 0.9340 & 0.9576 & 0.9301 & 0.9563 & 0.9344 & 0.9161 & \textbf{0.9608}\\
JP2K  & 0.7592 & 0.5097 & 0.8706 & 0.9650 & 0.9586 & 0.9706 & 0.9266 & 0.9614 & 0.9631 & 0.9396 & \textbf{0.9791}\\
JGTE  & 0.5604 & 0.4494 & 0.5191 & 0.8765 & 0.9297 & 0.9004 & 0.8739 & \textbf{0.9304} & 0.8926 & 0.8471 & 0.9057\\
J2TE  & 0.7003 & 0.1405 & 0.4322 & 0.8951 & \textbf{0.9034} & 0.8973 & 0.8139 & 0.9003 & 0.8311 & 0.7752 & 0.9016\\
NEPN  & 0.3111 & 0.2163 & 0.1230 & 0.4937 & 0.7238 & 0.5688 & 0.7011 & 0.6350 & 0.5266 & 0.7438 & \textbf{0.7563}\\
Block & 0.2659 & 0.3767 & 0.4059 & 0.5424 & 0.3899 & 0.4174 & 0.2739 & 0.5956 & 0.4866 & 0.5039 & \textbf{0.6434}\\
MS    & 0.1852 & 0.0633 & 0.4596 & 0.2249 & 0.4016 & -0.0261 & 0.5150 & 0.3628 & 0.1053 & \textbf{0.6322} & 0.4849 \\
CTC   & 0.0182 & 0.0466 & 0.5401 & 0.5842 & 0.7637 & 0.5785 & 0.3572 & 0.6588 & \textbf{0.8501} & 0.5742 & 0.6620\\
CCS   & 0.2142 & -0.1390 & 0.5640 & 0.6170 & 0.8294 & 0.7176 & 0.5632 & 0.7943 & 0.8302 & 0.7165 & \textbf{0.8441}\\
MGN   & 0.8777 & 0.5491 & 0.8810 & 0.9299 & 0.9392 & 0.9425 & 0.8945 & 0.9488 & 0.9239 & 0.9221 & \textbf{0.9535}\\
CN    & 0.4706 & 0.3740 & 0.6466 & 0.9365 & 0.9516 & 0.9538 & 0.8848 & \textbf{0.9616} & 0.9549 & 0.8903 & 0.9514\\
LCNI  & 0.8238 & 0.5053 & 0.6882 & 0.9674 & 0.9779 & 0.9713 & 0.9363 & 0.9792 & 0.9620 & 0.9491 & \textbf{0.9796}\\
ICQD  & 0.4883 & 0.8036 & 0.7965 & \textbf{0.9301} & 0.8597 & 0.9164 & 0.8767 & 0.9137 & 0.9098 & 0.8818 & 0.9175 \\
CHA   & 0.7470 & 0.6657 & 0.7950 & 0.8964 & \textbf{0.9269} & 0.9031 & 0.7955 & 0.9092 & 0.9086 & 0.8927 & 0.9243\\
SSR   & 0.7727 & 0.8273 & 0.8220 & 0.9538 & 0.9744 & \textbf{0.9754} & 0.9418 & 0.9600 & 0.9306 & 0.9272 & 0.9746\\ 
\midrule
Average & 0.5950 & 0.4725 & 0.7073 & 0.8293 & 0.8539 & 0.8234 & 0.7915 & 0.8588 & 0.8371 & 0.8444 & \textbf{0.8880}\\
\midrule
\rowcolor[gray]{0.9}
Hit Count & 0 & 0 & 0 & 1 & 2 & 1 & 0 & 3 & 2 & 1 & \textbf{14}\\ 
\bottomrule[1.1pt]
\end{tabular}}
\end{table*}

\begin{table*}[t!]
\centering
\caption{Leave-One-Distortion-Out Performance Comparison on the KADID-10K Dataset.}
\label{tabs6}
\resizebox{\textwidth}{!}{
\begin{tabular}{l|cccccccccc|c}
\toprule[1.1pt]
Dist. Type & BRISQUE & HOSA & WaDIQaM & DBCNN & MetalQA & HyperIQA & MUSIQ* & TReS* & Su \textit{et al.} & ARNIQA* & \textbf{Ours (R)}\\
\midrule
GB         & 0.8118  & 0.8522  & 0.8792  & 0.9549  & 0.9461  & 0.9464  & 0.9575 & 0.9568 & 0.9596 & 0.9454 & \textbf{0.9679}\\
LB         & 0.6738  & 0.7152  & 0.7299  & 0.9037  & 0.9168  & 0.9221  & 0.9213 & 0.9260 & 0.9241 & 0.9098 & \textbf{0.9566}\\
MB         & 0.4226  & 0.6515  & 0.7304  & 0.9116  & 0.9262  & 0.9340  & 0.9505 & 0.9333 & 0.9037 & 0.9448 & \textbf{0.9674}\\
CD         & 0.5440  & 0.7272  & 0.8325  & 0.8873  & 0.8917  & 0.9187  & 0.8425 & 0.8813 & 0.8966 & 0.7867 & \textbf{0.9288}\\
CS         & -0.1821 & 0.0495  & 0.4209  & 0.7116  & 0.7850  & 0.7835  & 0.6615 & \textbf{0.8462} & 0.7257 & 0.7078 & 0.7219\\
CQ         & 0.6670  & 0.6617  & 0.8055  & 0.8475  & 0.7170  & 0.8623  & 0.7724 & \textbf{0.8966} & 0.8725 & 0.8019 & 0.8439\\
CSA1       & 0.0706  & 0.2158  & 0.1479  & 0.3248  & 0.3039  & \textbf{0.4956}  & 0.4657 & 0.3867 & 0.3810 & 0.1545 & 0.4891\\
CSA2       & 0.3746  & 0.8408  & 0.8358  & 0.9128  & 0.9310  & 0.9396  & 0.9141 & 0.9213 & 0.9153 & 0.9000 & \textbf{0.9425}\\
JP2K       & 0.5159  & 0.6078  & 0.5387  & \textbf{0.9504}  & 0.9452  & 0.9178  & 0.9353 & 0.9277 & 0.9297 & 0.8977 & 0.9308\\
JPEG       & 0.7821  & 0.5823  & 0.5298  & 0.9122  & 0.9115  & 0.9181  & 0.8980 & 0.9363 & 0.9286 & 0.9062 & \textbf{0.9417}\\
WN         & 0.7080  & 0.6796  & 0.8966  & 0.9413  & 0.9047  & 0.9442  & 0.9291 & 0.9475 & 0.9549 & 0.9253 & \textbf{0.9561}\\
WNCC       & 0.7182  & 0.7445  & 0.9247  & 0.9631  & 0.9303  & 0.9646  & 0.9537 & 0.9678 & 0.9704 & 0.9497 & \textbf{0.9712}\\
IN         & -0.5425 & 0.2535  & 0.8142  & 0.8277  & 0.8673  & 0.8825  & 0.7985 & 0.8889 & 0.7369 & 0.8354 & \textbf{0.9116}\\
MN         & 0.6741  & 0.7757  & 0.8841  & 0.9228  & 0.9247  & 0.9638  & 0.9424 & 0.9625 & \textbf{0.9644} & 0.9422 & \textbf{0.9644}\\
Denoise    & 0.2213  & 0.2466  & 0.7648  & 0.8997  & 0.8985  & 0.9183  & 0.8429 & 0.9433 & 0.9353 & 0.9268 & \textbf{0.9557}\\
Brighten   & 0.5754  & 0.7525  & 0.6845  & 0.9072  & 0.7827  & 0.8327  & 0.8787 & 0.8716 & 0.8653 & 0.8719 & \textbf{0.9084}\\
Darken     & 0.4050  & 0.7436  & 0.2715  & 0.8029  & 0.6219  & 0.7114  & 0.7547 & 0.6685 & 0.8241 & 0.6871 & \textbf{0.8274}\\
MS         & 0.1441  & 0.5907  & 0.3475  & 0.6534  & 0.5555  & 0.6894  & 0.6169 & 0.6890 & 0.7105 & 0.6973 & \textbf{0.7967}\\
Jitter     & 0.6719  & 0.3907  & 0.7781  & 0.8839  & 0.9278  & 0.8900  & \textbf{0.9349} & 0.8616 & 0.8687 & 0.9287 & 0.9341 \\
NEP        & 0.1911  & 0.4607  & 0.3478  & 0.4214  & 0.4184  & 0.4373  & 0.5556 & 0.5206 & 0.5689 & \textbf{0.5993} & 0.5665 \\
Pixelate   & 0.6477  & 0.7021  & 0.6998  & 0.8610  & 0.8090  & 0.8688  & 0.8891 & 0.8871 & 0.8547 & 0.7872 & \textbf{0.8907}\\
Quantization & 0.7135 & 0.6811  & 0.7345  & 0.8199  & \textbf{0.8770}  & 0.8702  & 0.8463 & 0.8737 & 0.8424 & 0.8041 & 0.8612 \\
CB         & 0.0673  & 0.3879  & 0.1602  & 0.4014  & 0.5132  & 0.4539  & 0.4648 & 0.4674 & 0.4761 & 0.5991 & \textbf{0.6166}\\
HS         & 0.3611  & 0.2302  & 0.5581  & 0.9016  & 0.4374  & 0.8978  & 0.7859 & 0.9028 & 0.7730 & 0.8815 & \textbf{0.9220}\\
CC         & 0.1048  & 0.4521  & 0.4214  & \textbf{0.7138}  & 0.4377  & 0.5428  & 0.5461 & 0.5344 & 0.5603 & 0.5492 & 0.6292 \\
\midrule
Average & 0.4136 & 0.5598 & 0.6295 & 0.8095 & 0.7672 & 0.8202 & 0.8024 & 0.8240 & 0.8137 & 0.7976 & \textbf{0.8561} \\
\midrule
\rowcolor[gray]{0.9}
Hit Count & 0 & 0 & 0 & 2 & 1 & 1 & 1 & 2 & 1 & 1 & \textbf{17} \\ 
\bottomrule[1.1pt]
\end{tabular}}
\end{table*}

\begin{figure*}[t!]
\centering
\includegraphics[width=0.95\textwidth]{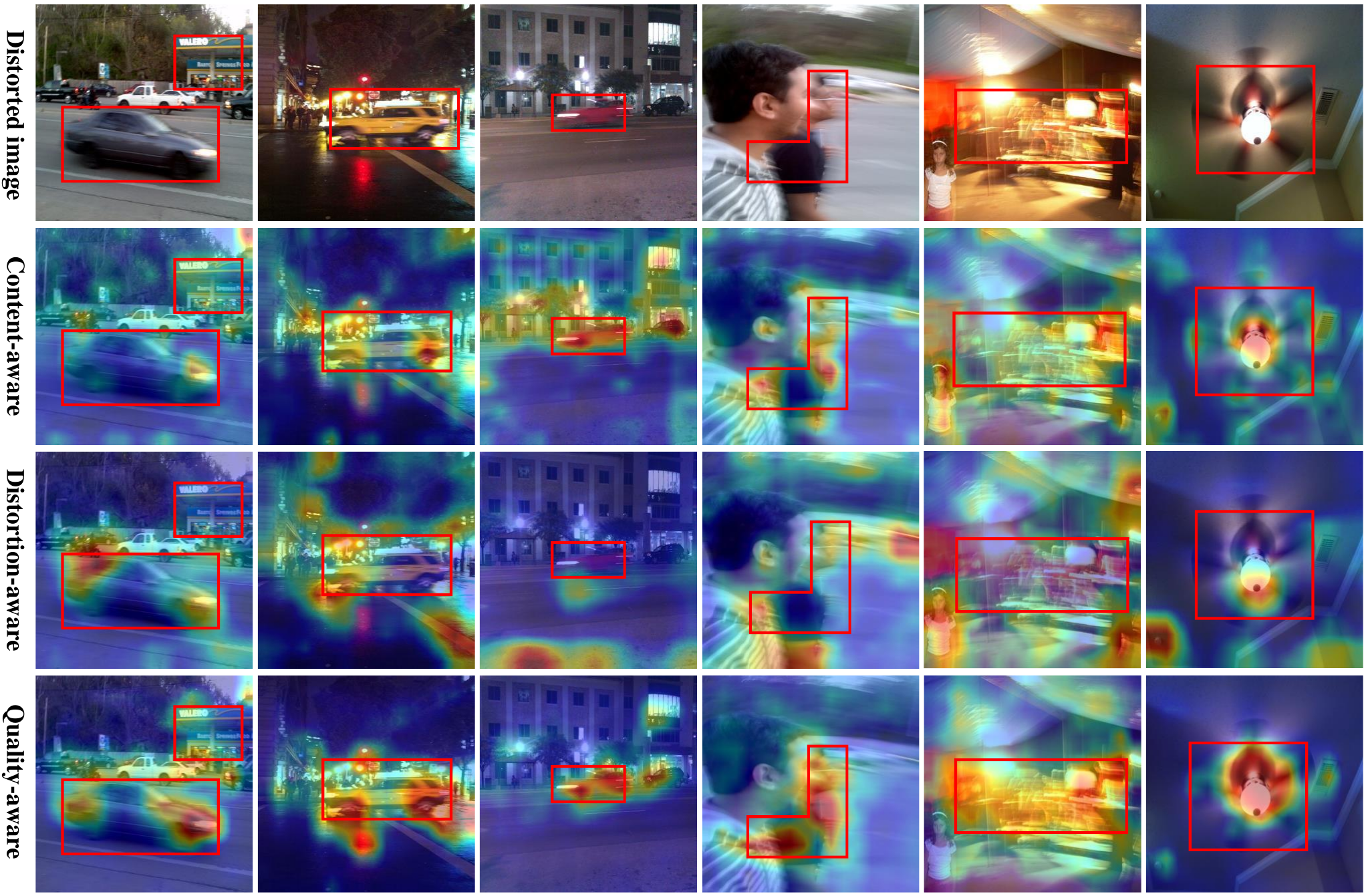}
\caption{Attention maps of content-aware, distortion-aware, and quality-aware features from CoDI-IQA.}
\label{fig9}
\end{figure*}
\subsection{Leave-One-Distortion-Out Validation}
Since the DAE within CoDI-IQA is pre-trained to learn the image distortion manifold, we wonder whether the representations still retain sensitivity to distortion information after high-order interaction through PPIMs. To validate this property, we conduct leave-one-distortion-out experiments on the TID2013 \cite{TID} and KADID-10K \cite{KADID} datasets. Following \cite{su2023DisManifold}, we iteratively select one distortion type for testing and use the remaining types for training, in order to evaluate the generalizability of the proposed method to unseen distortions. CoDI-IQA is compared with BRISQUE \cite{mittal2012BRISQUE},  HOSA \cite{xu2016HOSA}, WaDIQaM \cite{bosse2018WaDIQaM}, DBCNN \cite{zhang2020DBCNN}, MetalQA \cite{zhu2020metaiqa}, HyperIQA \cite{su2020HyperIQA}, MUSIQ \cite{ke2021musiq}, TReS \cite{golestaneh2022TReS}, Su \textit{et al.} \cite{su2023DisManifold}, and ARNIQA \cite{agnolucci2024ARNIQA}. From the main manuscript, we can see that the ResNet50 version of CoDI-IQA is already effective enough on synthetic datasets. Therefore, the Swin-based model is not included in these experiments. The SRCC results are reported in Table \ref{tabs5} and Table \ref{tabs6}, with the first column indicating the distortion type held out for testing. The last row in each table summarizes total hit counts of each method achieved the best performance across all distortion types. 

It can be observed that CoDI-IQA achieves the best performance in 31 out of 49 cases. It also improves the average SRCC to 0.8880 (+3.4\%) on TID2013 and 0.8561 (+3.9\%) on KADID-10K. These results clearly show that the proposed method has a significant advantage in recognizing unseen distortions. Although Su \textit{et al.} \cite{su2023DisManifold} and ARNIQA \cite{agnolucci2024ARNIQA} respectively adopt supervised and self-supervised learning to learn the distortion manifold while ignoring image content, their overall performance appears ordinary. This evidence confirms that proper interaction modeling can not only retain sensitivity to distortion information but also better reflect its combined influence with content on perceived quality. Moreover, CoDI-IQA achieves superior performance on distortions such as local block-wise artifacts (Block) in TID2013 and color block (CB) in KADID, which can corrupt high-level semantic information. This is because it is capable of capturing the hierarchical impact of such distortions on semantic meaning through its hierarchical interaction mechanism. However, the proposed method is not always the best. One plausible explanation is that distortions such as contrast change (CTC) impact the entire image uniformly without introducing significant pronounced interplay with content. As a result, CoDI-IQA performs inferior on these distortions. In summary, the consistent performance on both known and unseen distortions highlights the robustness and generalizability of the proposed method.

\subsection{More Detailed Visualization}
In the main manuscript, we only visualize the attention maps of the final quality-aware features from CoDI-IQA due to space limitations. Here, we provide detailed visualizations to more clearly show how PPIM models the interactions between content and distortions. Specifically, the attention maps of content-aware and distortion-aware features are included for comparison. As shown in Fig. \ref{fig9}, the content-aware features tend to highlight semantically meaningful regions, regardless of the severity of distortion. And the distortion-aware features localize areas that show obvious degradation, such as motion blur or overexposure. Neither of these two types of features alone can properly represent the regions that are most relevant to overall perceptual quality. For instance, the content-aware features of the first image do not adequately capture the moving car. The distortion-aware features in the fifth column overlook the most severely degraded region. In contrast, the final quality-aware features concentrate on regions where severe distortions are intertwined with important content. This is because the proposed PPIM leverages distortion location information to guide interaction modeling in a content-adaptive manner, which helps the model pinpoint the true areas of quality degradation. Consequently, CoDI-IQA is able to construct quality perception rules consistent with human visual perception.
\begin{table}[h]
\centering
\caption{Performance Comparison on the NNID Dataset. \\ Rows in Gray Denote Methods Designed for NTIs.}
\label{tabs7}
\resizebox{0.95\columnwidth}{!}{
\begin{tabular}{lcccc}
\toprule[1.1pt]
Methods& SRCC & PLCC & KRCC & \phantom{\textdownarrow}RMSE \textdownarrow \\
\midrule
BRISQUE \cite{mittal2012BRISQUE}   & 0.7365 & 0.7452 & 0.5352 & 0.1132 \\
HOSA \cite{xu2016HOSA}     & 0.5484 & 0.5487 & 0.3806 & 0.1416 \\
WaDIQaM \cite{bosse2018WaDIQaM}  & 0.8272 & 0.8229 & 0.6213 & 0.0954 \\
DBCNN \cite{zhang2020DBCNN}     & 0.8938 & 0.8958 & 0.6953 & 0.0849 \\
TOPIQ* \cite{chen2024topiq}    & 0.9360 & 0.9345 & 0.7763 & 0.0646 \\
\rowcolor[gray]{0.9}
BNBT \cite{xiang2019NNID}      & 0.8769 & 0.6822 & 0.8784 & 0.1061 \\
\rowcolor[gray]{0.9}
PCSNet \cite{li2025PCSNet}   & 0.9193 & 0.9160 & 0.7516 & 0.0713 \\
\midrule
\textbf{Ours (R)} & \underline{0.9396} & \underline{0.9367} & \underline{0.7820} & \underline{0.0604} \\
\textbf{Ours (S)} & \textbf{0.9500} & \textbf{0.9499} & \textbf{0.8037} & \textbf{0.0564} \\
\bottomrule[1.1pt]
\end{tabular}}
\end{table}
\subsection{Night-Time Image Quality Assessment}
Most of the methods that are introduced in the main manuscript are general-purpose NR-IQA methods. The proposed CoDI-IQA is also one of them. However, general-purpose methods may not perform well under challenging conditions such as night-time scenario. To investigate the practical applicability of the method, we further evaluate the effectiveness of CoDI-IQA on night-time images (NTIs). A commonly used natural NTI dataset is NNID \cite{xiang2019NNID}, which contains 2,240 NTIs with 448 distinct image contents. These images were captured using three different photographic devices in real-world scenarios, and are accompanied by corresponding subjective quality scores. In our experiments, all images from NNID are resized to 512×512 for training and evaluation. In addition to SRCC and PLCC, Kendall rank order correlation coefficient (KRCC) and root mean square error (RMSE) are also employed as evaluation criteria. We compare CoDI-IQA with several general-purpose NR-IQA methods and PCSNet \cite{li2025PCSNet}, which is specifically designed for NTIs. The media results are listed in Table \ref{tabs7}. It can be observed that the two variants of CoDI-IQA consistently achieve top-2 performance across all evaluation metrics, even though they are not specifically tailored for NTIs. We attribute this outcome to the fact that CoDI-IQA learns to model how the interactions between content and distortion manifest differently depending on the underlying image characteristics, which enables the model to adapt to challenging scenes such as night-time images. These results highlight the practical applicability of the proposed method.
\begin{table}[t]
\centering
\caption{Performance Comparison on Generic Face IQA Datasets. \\ Rows in Gray Denote BFIQA Methods \\ And Those in Light Blue Gray for GFIQA Methods}
\label{tabs8}
\resizebox{\columnwidth}{!}{
\begin{tabular}{lcccc}
\toprule[1.1pt]
\multirow{2}{*}{\centering Methods\vspace{-5pt}} & \multicolumn{2}{c}{GFIQA-20K} & \multicolumn{2}{c}{CGFIQA-40K} \\
\cmidrule(lr){2-3} \cmidrule(lr){4-5}
 & SRCC & PLCC &  SRCC & PLCC \\
\midrule
HyperIQA \cite{su2020HyperIQA}       & 0.967 & 0.966 & 0.973 & 0.972 \\
MetaIQA \cite{zhu2020metaiqa}      & 0.953 & 0.954 & 0.946 & 0.947 \\
MUSIQ \cite{ke2021musiq}        & 0.952 & 0.950 & 0.974 & 0.975 \\
TReS  \cite{golestaneh2022TReS}         & 0.955 & 0.951 & 0.982 & 0.982 \\
CONTRIQUE \cite{madhusudana2022CONTRIQUE}     & 0.947 & 0.946 & 0.980 & 0.979 \\
Re-IQA \cite{saha2023ReIQA}      & 0.945 & 0.944 & 0.980 & 0.980 \\
TOPIQ* \cite{chen2024topiq}          & 0.965 & 0.965 & 0.984 & 0.985 \\
\rowcolor[gray]{0.9}
ArcFace \cite{deng2019arcface}       & 0.951 & 0.951 & 0.972 & 0.972 \\
\rowcolor[gray]{0.9}
MegaFace \cite{meng2021magface}      & 0.953 & 0.952 & 0.973 & 0.973 \\
\rowcolor[gray]{0.9}
CR-FIQA \cite{boutros2023cr}      & 0.960 & 0.959 & 0.974 & 0.973 \\
\rowcolor{LightBlue}
IFQA \cite{jo2023ifqa}          & 0.960 & 0.960 & 0.980 & 0.979 \\
\rowcolor{LightBlue}
StyleGAN-IQA \cite{GFIQA}   & 0.968 & 0.967 & 0.982 & 0.982 \\
\rowcolor{LightBlue}
DSL-FIQA \cite{CGFIQA}       & \textbf{0.975} & \textbf{0.974} & \textbf{0.988} & \underline{0.987} \\
\midrule
\textbf{Ours (R)}    & 0.967 & \underline{0.968} & \underline{0.986} & 0.986 \\
\textbf{Ours (S)}    & \underline{0.974} & \textbf{0.974} & \textbf{0.988} & \textbf{0.988} \\
\bottomrule[1.1pt]
\end{tabular}}
\end{table}
\subsection{Generic Face Image Quality Assessment}
Unlike natural images, face images are inherently more complex due to subtle visual features and expressions, which significantly influence the perceived image quality. Existing general-purpose NR-IQA methods often perform sub-optimally on face images, as they fail to capture the distinct characteristics and subtle variations inherent in human faces. To further enrich the experimental results and demonstrate the real-world applicability of the proposed method, we conduct experiments on two generic face IQA datasets. The GFIQA-20K \cite{GFIQA} dataset comprises 20,000 face images, which are split into 14,000 for training, 2,000 for validation, and 4,000 for testing. The CGFIQA-40K \cite{CGFIQA} dataset provides a more extensive collection of 39,312 images, with 27,518 used for training, 3,931 for validation, and 7,863 for testing. CoDI-IQA is compared against seven general-purpose NR-IQA methods, three biometric face image quality assessment (BFIQA) methods \cite{deng2019arcface, meng2021magface, boutros2023cr}, and three generic face image quality assessment (GFIQA) methods \cite{jo2023ifqa, GFIQA, CGFIQA}. The SRCC and PLCC results are summarized in Table \ref{tabs8}. 

DSL-FIQA \cite{CGFIQA} is the current SOTA method for GFIQA. It combines dual-set degradation representation learning with a landmark-guided transformer architecture to focus on salient facial regions. Without any face-specific design and only replacing the CAE, our CoDI-IQA chieves performance on par with DSL-FIQA. The superior performance can be attributed to the following two reasons. First, the data distributions of the two datasets are well balanced, which has a significant positive impact on model performance. As a result, all methods achieve reasonably good performance under this setting. Second, face image quality heavily relies on content-dependent predictions. CoDI-IQA is highly compatible with this property, as the fine interaction step within PPIM effectively captures local interaction patterns while preserving semantic integrity. Notably, our intention is solely to validate the practical applicability of the proposed method across different scenarios. For more details about GFIQA, please refer to \cite{GFIQA,CGFIQA}.
\end{document}